\documentclass[a4paper]{article}
\pdfoutput=1
\usepackage[utf8]{inputenc} 
\usepackage[T1]{fontenc}    
\usepackage{lmodern}
\usepackage{hyperref}       
\usepackage{url}            
\usepackage{booktabs}       
\usepackage{amsfonts}       
\usepackage{nicefrac}       
\usepackage{microtype}      

\usepackage{graphicx}
\usepackage{pdfpages}
\usepackage{multirow}
\usepackage{caption}
\usepackage{subcaption}
\usepackage[a4paper,margin=1in,footskip=0.25in]{geometry}
\usepackage{amsmath, amssymb}
\usepackage{enumitem}

\usepackage[sort, numbers]{natbib}

\usepackage{epstopdf}

\newif\ifOmitDiscussion
\OmitDiscussiontrue

\newif\ifReview
\Reviewtrue

\newif\ifSuppressMemo
\ifSuppressMemo
\newcommand{\memo}[1]{}
\else
\usepackage{color}
\newcommand{\memo}[1]{{\bf \textcolor{red}{[#1]}}}

\fi

\usepackage{nkj,nkj_e}

\newcommand{\Obj}{g}
\newcommand{\ObjInt}{G}

\newcommand{\method}{MALADE}

\title{Robustifying Models Against Adversarial Attacks \\by Langevin Dynamics}
\author{
 Vignesh Srinivasan$^{1}$, Arturo Marban$^{1,2,3}$, Klaus-Robert M{\"u}ller$^{2,3,4,5,\thanks{Asterisks indicate corresponding author}}$\\
{  {Wojciech Samek$^{1,3,\footnotemark[1]}$
and Shinichi Nakajima$^{2,3,6,\footnotemark[1]}$} }\\
$^1$Fraunhofer HHI, $^2$TU Berlin, $^3$Berlin Big Data Center, \\
 $^4$Korea University, $^5$MPI for Informatics, $^6$RIKEN AIP\\ 
\texttt{\{vignesh.srinivasan,wojciech.samek\}@hhi.fraunhofer.de}\\
\texttt{\{arturo.marban,klaus-robert.mueller,nakajima\}@tu-berlin.de }\\
}
\begin{document}



\date{}
\maketitle


\begin{abstract}

Adversarial attacks on deep learning models have compromised their performance considerably.
As remedies,
a lot of defense methods were proposed, 
which however, have been circumvented by newer attacking strategies.
In the midst of this ensuing arms race,
the problem of robustness against adversarial attacks still remains unsolved. 
This paper proposes a novel, simple yet effective defense strategy
where adversarial samples are relaxed onto the underlying manifold of the (unknown) target class distribution. Specifically, our algorithm
drives off-manifold adversarial samples towards high density regions of the data generating distribution of the target class by the Metroplis-adjusted Langevin algorithm (MALA)
with \emph{perceptual boundary taken into account}.
Although the motivation is similar to projection methods, e.g., Defense-GAN,
our algorithm, called MALA for DEfense (\method{}), is equipped with significant dispersion---projection is distributed broadly, and therefore any whitebox attack cannot accurately align the input so that the \method{} moves it to a targeted untrained spot where the model predicts a wrong label.
In our experiments, 
\method{} exhibited state-of-the-art performance against various elaborate attacking strategies.

\end{abstract}

\section{Introduction}
\label{sec:Introduction}
Deep neural networks (DNNs) \citep{Krizhevsky,lenet,googlenet,vgg,resnet}
have shown excellent performance in many applications,
while
they are known to be susceptible to adversarial attacks, i.e.,
examples crafted intentionally by adding slight noise to the input \citep{Goodfellow, Papernotb, szegedy, Nguyena, Eykholt,athalye3d}.
These two aspects are considered to be two sides of the same coin:
deep structure induces complex interactions between weights of different layers, which provides flexibility in expressing complex input-output relation with relatively small degrees of freedom, while it can make the output function unpredictable in \emph{spots} where training samples exist sparsely.
If adversarial attackers manage to find such spots in the input space close to a \emph{real} sample, they can manipulate the behavior of classification, which can lead to a critical risk of security in applications, e.g., self-driving cars, for which high reliability is required.

Different types of defense strategies 
were proposed,
including 
\emph{adversarial training} \citep{Floriantra, distill, Strauss2018, Gu, Lamb, madry2017towards, kannan2018adversarial, liu2018adv, xie2018feature} which 
incorporates adversarial samples in the training phase, 
\emph{projection methods} \citep{schott2018robust, song2017pixeldefend, Pouya, Ilyas2017, Lamb, shen2017ape}
which denoise adversarial samples by projecting them onto the data manifold,
and \emph{preprocessing methods} \citep{chuan_fb, Liao, Meng,xie2017mitigating}
which try to destroy elaborate spatial coherence hidden in adversarial samples. 
Although those defense strategies were shown to be robust against the attacking strategies that had been proposed before, most of them 
have been circumvented by newer attacking strategies.
Another type of approaches, called \emph{certification-based methods}
\cite{wong2017provable, wong2018scaling, raghunathan2018certified, dvijotham2018dual, xie2018feature},
minimize (bounds of) the worst case loss over a defined range of perturbations,
and provide theoretical guarantees on robustness against any kind of attacks.
However, the guarantee holds only for small perturbations,
and the performance of those methods against existing attacks are typically inferior to the state-of-the-art.
Thus, the problem of robustness against adversarial attacks still remains unsolved.

\begin{figure}[t]
\centering
\includegraphics[width=0.75\linewidth]{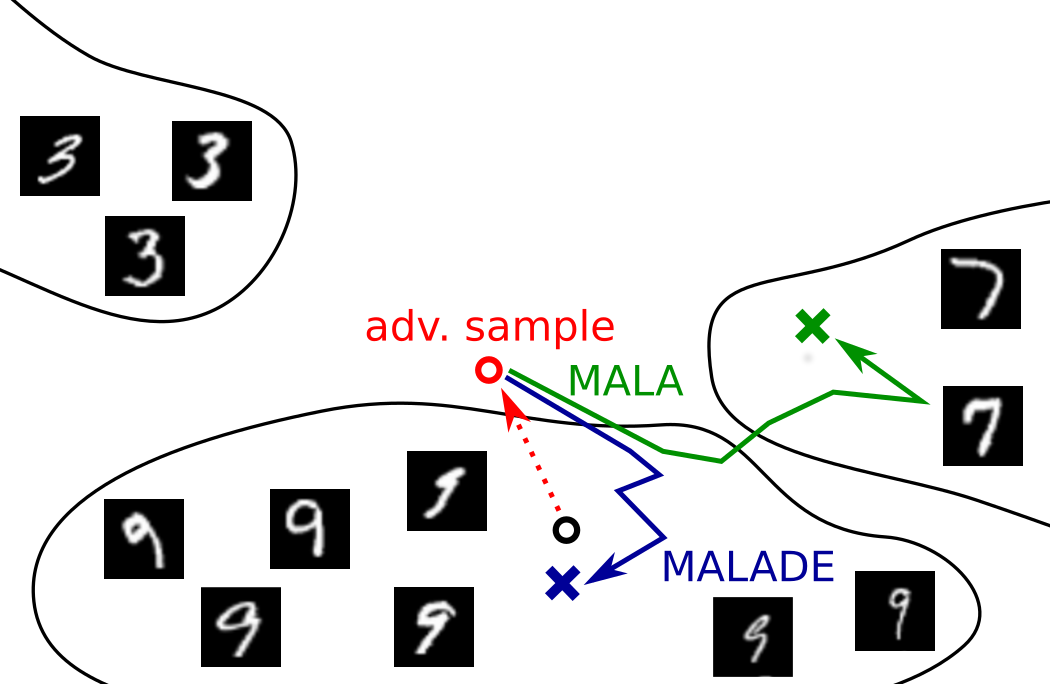}
\caption{An adversarial sample (red circle) is created by moving the data point (black circle) away from the data manifold, here the manifold of images of digit "9". In this low density area, the DNN is not well trained and thus misclassifies the adversarial sample. Unsupervised sampling techniques such as MALA (green line) project the data point back to high density areas, however, not necessarily to the manifold of the original class. The proposed \method{} (blue line) takes into account of class information and thus projects the adversarial sample back to the manifold of images of digit "9".} 
\label{fig:cartoon_fig}    
\hspace{-5mm}
\end{figure}

In this paper,
we propose a novel defense strategy, which drives adversarial samples
towards high density regions of the data distribution. 
Fig.~\ref{fig:cartoon_fig} explains the idea of our approach.
Assume that an attacker created 
an adversarial sample (red circle) by moving an original sample (black circle) to an \emph{untrained spot} where the target classifier gives a wrong prediction.
We can assume that the spot is in a low density area of the training data, i.e., off the data manifold, where
the classifier is not well trained, but still close to the original high density area
so that the adversarial pattern is imperceptible to a human.
Our approach is to \emph{relax} the adversarial sample by Metropolis-adjusted Langevin algorithm (MALA) \citep{mala_a, mala_b}, an efficient Markov chain Monte Carlo (MCMC) sampling method. 

MALA requires the gradient of the energy function, 
which corresponds to the gradient $ \bfnabla_{\bfx}\log p(\bfx)$ of the log probability, a.k.a., the score function,
of the input distribution. 
However, 
 naively applying MALA would have an apparent drawback: if there exist high density regions (clusters) close to each other but not sharing the same label, MALA could drive a sample into another cluster (see green line in Fig.\ref{fig:cartoon_fig}), which degrades the classification accuracy. 
To overcome this drawback, we replace the (marginal) distribution $p(\bfx)$ 
with the conditional distribution $p(\bfx|\bfy)$ given label $\bfy$.
More specifically, 
our novel defense method, called MALA for DEfense (\method{}),
relaxes the adversarial sample based on the conditional gradient $ \bfnabla_{\bfx}\log p(\bfx| \bfy)$
by using a novel estimator for the conditional gradient
\emph{without knowing the label $\bfy$ of the test sample}.
Thus, \method{} drives the adversarial sample towards high density regions of the data generating distribution for the original class (see blue line in Fig.\ \ref{fig:cartoon_fig}),
where the classifier is well trained to predict the correct label.

Our proposed \method{} can be seen as one of the projection methods,
most of which have been circumvented by recent attacking methods.
However, \method{} has two essential differences from the previous projection methods:
\begin{description}
\item [Perceptual boundary taken into account]
    All previous projection methods, including Defense-GAN \citep{Pouya}, PixelDefend \citep{song2017pixeldefend},
    and others \citep{Ilyas2017,shen2017ape}, pull the input sample into the closest point on the data manifold without the label information into account.
    On the other hand, \method{} is designed to drive the input sample into the data manifold of the original class.
\item [Significant dispersion]
     Most projection methods try to pull the adversarial sample back to the original point (so that the adversarial pattern is removed).
     On the other hand, \method{} drives the input sample to anywhere (randomly) in the closest cluster having the original label.
     In this sense, \method{} has much larger inherent randomness than the previous projection methods,
\end{description}
Significant dispersion effectively prevents any whitebox attack from  aligning the input so that the \method{} \emph{stably} moves it to a targeted untrained spot,
while the awareness of the perceptual boundary keeps the sampling sequence 
within the cluster with the correct label.
Furthermore, it is straightforward to combine \method{} with the state-of-the-art adversarial training method---an advantage to be a projection method---which further boosts the performance.

Our experiments show that \method{} performs comparably to the state-of-the-art methods for older attacks (which had been proposed before the baseline defense methods were proposed) with low perturbation intensity, 
while it significantly outperforms the state-of-the-art methods for newer attacks or the older attacks with high perturbation intensity.
In our experiments, we paid careful attention to the fairness of evaluation, following the recommendation by 
recent papers \citep{engstrom2018evaluating, athalye2018obfuscated, uesato2018adversarial, carlini2019evaluating},
where 
inappropriate evaluation of defense strategies in literature
was reported.
The authors pointed out that many defense methods were proposed showing good performance against weak versions of attacking strategies,
and argued that, when a new defense method is proposed, it should be evaluated throughly
by appropriately choosing the hyperparameters of attacks,
and 
by elaborating most effective attacks against the proposed defense method.
We did our best in this regards,
and explored most effective combinations of existing attacking strategies so that the attack is considered to be strongest against our own method.

This paper is organized as follows.
We first summarize existing attacking and defense strategies
in Section \ref{sec:Existing}.
Then, we propose our method with a novel conditional gradient estimator
in Section~\ref{method}.
In Section~\ref{exp},
we evaluate our defense method against various attacking strategies,
and show advantages over the state-of-the-art defense methods.
Section~\ref{conclusion} concludes.

%
%
%
%

\section{Existing Methods}
\label{sec:Existing}

In this section, we introduce existing attacking and defense strategies.

\subsection{Attacking Strategies}
\label{sec:attack}

There are two scenarios considered in adversarial attacking.
The whitebox scenario assumes that the attacker has the full knowledge
on the target classification system,
including the architecture and the weights of the DNN and the respective defense strategy,%
%
\footnote{
As in \citep{athalye2018obfuscated},
we assume that the attacker cannot access to the random numbers (nor random seed) generated 
and used in stochastic defense processes.
}
while the blackbox scenario assumes that 
the attacker has access only to the classifier outputs.

\subsubsection{Whitebox Attacks (General)}
\label{sec:attackGeneral}


We first introduce representative whitebox attacks, which are effective against general classifiers with or without defense strategy.


\paragraph{Fast Gradient Sign Method (FGSM) \citep{Goodfellow}}
This method is one of the simplest and fastest attacking algorithms. Given an original sample $\bfx \in \mathbb{R}^L$ and its corresponding label $\bfy \in \{0, 1\}^K$ in the 1-of-$K$ expression, FGSM 
creates an adversarial sample by
\begin{equation}
\bfx' = \bfx + \varepsilon \cdot \mathrm{sign}(\bfnabla_{\bfx} J (\bfx,\bfy)),
\label{eq:FGSM}
\end{equation}
where $J (\bfx,\bfy) = - \sum_{k=1}^K y_k \log \widehat{y}_k(\bfx)$ is the cross entropy loss of the classifier output $\widehat{\bfy}  \in [0, 1]^K$ for the true label $\bfy$, and $\varepsilon$ is the amplitude of adversarial pattern.
Eq.\eqref{eq:FGSM} effectively reduces the classifier output
for the true label,
keeping the adversarial sample within the $\varepsilon$-ball around the original sample in terms of the $L_\infty$-norm.

\paragraph{Projected Gradient Descent (PGD) \citep{madry2017towards}}
PGD, a.k.a., the basic iterative method, solves the following problem:
\begin{align}
& \min_{\bfx'} J (\bfx',\bfy) 
\notag\\
& \textrm{s.t.} \quad  \|\bfx' - \bfx\|_p \leq \varepsilon,
\quad  \bfx'\in [0,1]^L,
\label{eq:PGD}
\end{align}
where $\|\cdot\|_p$ denotes the $L_p$-norm.
This problem can be solved by iterating FGSM \eqref{eq:FGSM}
as a gradient step and projecting the sample
to the allowed set of perturbations. 
PGD can be seen as an enhanced version of FGSM,
and is considered as the strongest first-order attack.

\paragraph{Momentum Iterative fast gradient sign Method (MIM) \citep{dong2018mim}}
MIM improves the convergence of the PGD problem \eqref{eq:PGD} by using the momentum, i.e.,
it iterates
\begin{equation}
\bfx' = \bfx + \varepsilon \cdot \mathrm{sign}(\bfg'),
\label{eq:MIM}
\end{equation}
where 
\begin{align}
\bfg' = \mu \cdot \bfg  + \frac{\bfnabla_{\bfx} J (\bfx,\bfy)}{\| \bfnabla_{\bfx} J (\bfx,\bfy) \|_{p}},
\notag
\end{align}
and the projection step.

\paragraph{Carlini-Wagner (CW)  \citep{Carlinib}}
The CW attack optimizes the adversarial pattern $\bftau \in \mathbb{R}^L$ by solving
\begin{align}
\min_{\bftau}&  \| \bftau\|_{p} + c \cdot F (\bfx + \bftau, \bfy) 
\notag\\
& \textrm{s.t.} \qquad  \bfx + \bftau \in [0,1]^L,
\label{eq:CW}
\end{align}
where 
\begin{align}
F(\bfx + \bftau, \bfy)
&= \max\{0, \log \widehat{y}_{k^*} (\bfx + \bftau) 
\notag \\
& \qquad \qquad
- \max_{k \ne k^*}\log \widehat{y}_{k}(\bfx + \bftau) + \iota\}.
\notag
\end{align}
Here, $c$ is a trade-off parameter balancing the pattern intensity and the adversariality,
$k^*$ is the true label id, i.e., $y_{k^*} = 1$,
and
$\iota$ is a margin for the sample to be adversarial.


\paragraph{Elastic-net Attack to Deep neural networks (EAD) \citep{chen2017ead, sharma2017ead}}
EAD is a modification of the CW attack 
where the $L_p$ regularizer is replaced with the 
elastic-net regularizer, the sum of $l_1$- and $l_2$-norms: 
\begin{align}
\min_{\bftau}&  \beta \| \bftau\|_{1} + \| \bftau\|_{2} + c \cdot 
F (\bfx + \bftau, \bfy) 
\notag\\
& \textrm{s.t.} \qquad  \bfx + \bftau \in [0,1]^L,
\label{eq:EAD}
\end{align}
where $\beta$ is a parameter controlling the sparsity.
This method creates sparse adversarial patterns,
and has shown to be very effective against the classifiers
that are protected against dense, e.g., $L_\infty$-bounded, adversarial patterns.


\subsubsection{Whitebox Attacks (Specialized)}
\label{sec:attacksForObfuscation}

Some attacking strategies target specific features of defense strategies,
and enhance general whitebox attacks (introduced in the previous subsection).
Here we introduce strategies that are considered to be effective 
for attacking our defense method, which will be proposed in Section~\ref{method}.

\paragraph{Reconstruction (R) Regularization \citep{frosst2018darccc}}
Some defense strategies are equipped with a denoising process,
where adversarial pattern is removed by projection, e.g., by an autoencoder.
The R regularization is designed to disarm those defense strategies
by finding a point which is unaffected by the denoising process.
In the original work, called R+PGD \citep{frosst2018darccc},
the sum of the reconstruction loss by the denoising process
and the (negative) cross entropy loss is minimized.


\paragraph{Back Pass Differentiable Approximation (BPDA) \citep{athalye2018obfuscated}}
Many defense strategies, explicitly or implicitly, rely on \emph{gradient obfuscation},
which prevents whitebox attackers from stably computing the gradient, e.g.,
by having non-differentiable layers or artificially inducing randomness.
BPDA simply replaces such layers with identity maps,
in order to stably estimate the gradient.
This method is effective when the replaced layer is the denoising process
that reconstructs the original input well, so that
$\bfnabla_{\bfx} \widehat{y}_k (r(\bfx')) \approx  \bfnabla_{\bfx} \widehat{y}_k(\bfx)$ 
holds.

\paragraph{Expectation over Transformation (EOT) \citep{athalye2018obfuscated}}
The EOT method estimates the gradient by averaging over multiple trails,
so that the randomness is averaged out. 
This method is effective against any stochastic defense methods.

\subsubsection{Blackbox Attacks} 

Blackbox attacks are, by definition, weaker than whitebox attacks.
Here we introduce a few of them.

\paragraph{Distillation Attack \citep{Papernot, Papernotb}}
One can apply any whitebox method after knowledge distillation, where a student network is trained from classifier outputs. 
Distillation attacks are, in principle, weaker than the corresponding whitebox attacks.

\paragraph{Boundary Attack \cite{brendel2017decision}}
This method first finds an initial adversarial point 
$\bfx'_{0}$ 
by random search.
Specifically, it adds uniform noise to the original sample
until the sample becomes adversarial, i.e., $\widehat{\bfy}(\bfx'_{0}) \ne \bfy$. 
Once a initial point is found, it iterates a random search step to reduce the norm of the adversarial pattern.
In each iteration, random exploration first finds the perceptual boundary on the sphere that is centered at the original sample $\bfx$, and contains the current point $\bfx'_{t}$.
Based on the boundary information,
the point is moved towards the original sample so that the 
distance to the original sample is reduced, i.e., $\|\bfx'_{t+1} - \bfx \|_p < \|\bfx'_{t} - \bfx \|_p$,
while the adversariality is kept, i.e., $\widehat{\bfy}(\bfx'_{t+1}) \ne \bfy$.
This process is iterated $T$ times.

\paragraph{Salt and Pepper Noise Attack \citep{hendrycks2018benchmarking} }

This method corrupts the original image with a fixed amount of impulsive salt and pepper noise.
The number of corrupted pixels is bounded by $\rho L$ for $0 < \rho \leq 1$. 
It was reported that the sharp discontinuity of
the salt and pepper noise enables strong adversarial attack,
bringing down the classification accuracy significantly \citep{hendrycks2018benchmarking}. 


\paragraph{Transfer Attack \cite{tramer2017space}}
Assume that the attacker has full access to another classifier which was trained for the same purpose (possibly equipped with some defense strategy) as the target classifier.
Then, the attacker can create adversarial samples by any whitebox attack against the known model,
and use them to attack the target classifier.
This attack can be considered as blackbox.

\subsection{Defense Strategies}
\label{defense_intro}

As mentioned in Section~\ref{sec:Introduction}, existing methods can be roughly classified into four categories. 

%
\paragraph{Adversarial Training}
In this strategy,
adversarial samples are generated by known attacking strategies, and added to the training data, 
in order to make the classifier robust against those attacks \citep{Papernotb, Gu, KurakinScale, Lamb, madry2017towards, kannan2018adversarial, liu2018adv, xie2018feature}.  
The method \citep{madry2017towards} proposed by Madry et al. (2017), which we refer to as "Madry" in this paper,
withstood many adversarial attacks,
and is 
considered to be the current state-of-the-art defense strategy that outperforms most of the other existing defense methods against most of the attacking methods. 
Adversarial Logit Pairing (ALP) \citep{kannan2018adversarial}
uses adversarial samples during training while performing logit pairing between adversarial samples and original samples thereby pulling the logits of the adversarial sample to be as similar as possible to the original. 
Although it was found \citep{engstrom2018evaluating} that ALP
is not as robust as what the original paper reported,
it is still considered to be one of the state-of-the-art methods.




Methods in this category,
typically
show higher robustness than methods in the other categories.
However, they have a risk of overfitting to the known attacking strategies \citep{chen2017ead, sharma2017ead}
and to intensity of perturbations \citep{madry2017towards}.
Our experiments in Section~\ref{exp} reveal this weakness of adversarial training---%
Madry
performs very well against 
FGSM \citep{Goodfellow}, PGD \citep{madry2017towards}, CW \citep{Carlinib}, and MIM \citep{dong2018mim},
while it was broken down by EAD \citep{chen2017ead, sharma2017ead},
which was proposed after Madry was proposed.

\paragraph{Projection Methods}


Generative models 
have been shown to be useful to reconstruct the original image (or to remove the adversarial patterns)
 from an adversarial sample \citep{Pouya,Ilyas2017,shen2017ape}. 
Defense-GAN \citep{Pouya} searches over the latent space of a generative adversarial network (GAN),
and provides a generated sample closest to the input as a reconstruction of the original image.
Since the generative model is trained to generate samples in the data manifold,
Defense-GAN effectively projects off-manifold sample onto the data manifold.
Similarly, PixelDefend \citep{song2017pixeldefend} employs PixelCNN to project the adversarial image back to the training distribution.
\citep{Ilyas2017} proposed an untrained generative network as a deep image prior to reconstruct the original image. 
The Analysis By Synthesis (ABS) method 
\citep{schott2018robust} uses a variational autoencoder to find the optimal latent vector maximizing the lower bound of the log likelihood of the given input to each of the class. 

Most existing strategies in this category have been rendered as ineffective against recent attacking strategies: Defense-GAN  and PixelDefend  have been broken down by 
BPDA \citep{athalye2018obfuscated}.

\begin{figure*}[t]
\centering
    \includegraphics[width=0.85\linewidth]{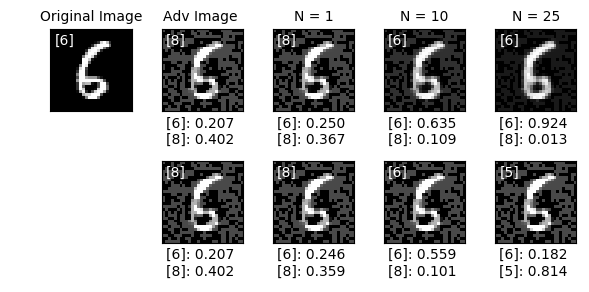}
\vspace{-2mm}
\caption{The top-left is the original image, from which the adversarial image (second column) was crafted.
The third to the fifth columns show the images after $N = 1, 10$, and $25$ steps
of \method{} (top-row) and of MALA with the marginal distribution  (bottom-row). 
Below each image, the prediction output $\widehat{y}_{k^*}$ for the original label $k^*=$"6"
and that for the label with the highest output, i.e., $k = \argmax_{k \ne k*} \widehat{y}_{k}$, are shown. 
In this example,
\method{},
trained with perceptional information,
drived the adversarial sample
towards the right cluster with the original label "6".
On the other hand,
although MALA successfully removed the adversarial pattern for "8", it brought the sample into a neighboring cluster with a wrong label "5".
}
\vspace{-2mm}
\label{fig:first_fig}      
\end{figure*}

\paragraph{Preprocessing Methods}
Preprocessing was known to be effective for adversarial defense.
Specifically,
it was reported that image transformations (e.g., bit depth reduction, JPEG compression and decompression, random padding) can destroy elaborate spatial coherence hidden in adversarial samples \citep{chuan_fb, xie2017mitigating, guo2017countering}.
However, 
\citep{athalye2018obfuscated} showed that this kind of defense strategies can be easily broken down 
by BPDA or EOT.
Autoencoders can also be used as a preprocessor to remove adversarial patterns \citep{Liao,Meng},
which however were broken down by CW \citep{carlini2017magnet}.

The methods in this category are generally seen as weaker than those in the other categories.

\paragraph{Certification-based Methods}
Certification-based methods employ robust optimization, and obtain \emph{provably robust} networks.
The idea is to train the classifier by minimizing (upper-bounds of) the worst-case loss over a defined range of perturbations, so that creating adversarial samples in the range is impossible.
To this end, one needs to solve a nested optimization problem, which consists of the inner optimization finding the worst case sample (or the strongest adversarial sample) and the outer optimization minimizing the worst-case loss.
Since the inner optimization is typically non-convex, different relaxations
\cite{wong2017provable,sinha2017certifying,raghunathan2018certified,dvijotham2018dual,wong2018scaling}
have been applied for scaling this approach.


The certification-based approach seems a promising direction
towards the end of the arms race---it might protect classifiers against any (known or unknown) attacking strategy in the future.
However, the existing methods still have limitations in many aspects, e.g., structure of networks, scalability, and the guaranteed range of the perturbation intensity.
Typically, the robustness is guaranteed only for small perturbations, e.g., 
$\varepsilon=0.1$ $L_{\infty}$-bound in MNIST \cite{wong2018scaling, raghunathan2018certified, dvijotham2018dual},
and no method in this category has shown comparable performance to the state-of-the-art.

\section{Proposed Method}
\label{method}

In this section, we propose our novel defense strategy,
which drives the input sample (if it lies in low density regions) towards high density regions.  We achieve this by using Langevin dynamics.

\paragraph{Metropolis-adjusted Langevin Algorithm (MALA)}
MALA
is an efficient Markov chain Monte Carlo (MCMC) sampling method
which uses the gradient of the energy (negative log-probability $E(\bfx) = - \log p(\bfx)$). 
Sampling is performed sequentially by
\begin{align}
\bfx_{t+1} = \bfx_{t} + \alpha \bfnabla_{\bfx} \log p(\bfx_{t}) + \bfkappa,
\label{eq:MalaDAE}
\end{align}
where $\alpha$ is the step size, and
$\bfkappa$ is random perturbation subject to $\mathcal{N}(\bfzero,\delta^{2}\bfI_L)$.
By appropriately controlling the step size $\alpha$ and the noise variance $\delta^{2}$,
the sequence is known to converge to the distribution $p(\bfx)$.%
\footnote{
For convergence, a rejection step after Eq.\eqref{eq:MalaDAE} is required.
However, it was observed that a variant, called MALA-approx \citep{Nguyen}, without the rejection step gives reasonable sequence for moderate step sizes.
We use MALA-approx in our proposed method.
}

We estimate the gradient, the second term in Eq.\eqref{eq:MalaDAE},
by using a denoising autoencoder (DAE) \citep{Vincent08}.
Let $\bfr: \mathbb{R}^{L} \mapsto \mathbb{R}^{L}$ be a function that minimizes
 \begin{align}
 \expect{ \|\bfr (\bfx + \bfnu) -  \bfx  \|^2  }{p'(\bfx) p'(\bfnu)},
 \label{eq:DAEObjective}
 \end{align}
 where $\expect{ \cdot}{p}$ denotes the expectation over the distribution $p$,
 $\bfx \in \mathbb{R}^L$ is a training sample subject to a distribution $p(\bfx)$, and $\bfnu \sim \mathcal{N}_L (\bfzero, \sigma^2 \bfI)$ is an $L$-dimensional artificial Gaussian noise with mean zero and variance $\sigma^2$.
 $p'(\cdot)$ denotes an empirical (training) distribution of the distribution $p(\cdot)$, namely,
 $\expect{g(\bfx)}{p'(\bfx)} = N^{-1}\sum_{n=1}^N g(\bfx^{(n)})$ where $\{\bfx^{(n)}\}_{n=1}^N$ are the training samples.
\begin{proposition}\citep{Alain14}
\label{prpt:DAEResidual}
Under the assumption that $\bfr(\bfx) = \bfx + o(1)$%
\footnote{
This assumption is not essential as we show in the proof in Appendix~\ref{sec:ProofSDAEScore}.
}, 
the minimizer of the DAE objective \eqref{eq:DAEObjective} satisfies
\begin{align}
\bfr(\bfx) - \bfx  = \sigma^2 \bfnabla_{\bfx} \log p(\bfx) 
+ o(\sigma^2),
\label{eq:ResidualScore}
\end{align}
as $\sigma^2 \to 0$.
\end{proposition}
Proposition~\ref{prpt:DAEResidual} states that 
a DAE trained with a small $\sigma^2$ can be used to estimate the gradient of the log probability.
In a blog \citep{HelsinkiAI}, it was proved that
the residual is proportional to the score function of the noisy input distribution for any $\sigma^2$, i.e.,
\begin{align}
\bfr(\bfx) - \bfx  = \sigma^2 \bfnabla_{\bfx}
\log \int  \mathcal{N}_L(\bfx; \bfx', \sigma^2 \bfI_L)p(\bfx') d\bfx'.
\label{eq:ResidualScoreConvoluted}
\end{align}

\paragraph{MALA for Defense (\method{})}

As discussed in Section~\ref{sec:Introduction},
MALA drives the input into high density regions but not necessarily to the cluster sharing the same label with the original image (see Fig.~\ref{fig:cartoon_fig}).
To overcome this drawback, we propose MALA for defence (\method{}),
where samples are relaxed based on the \emph{conditional} training distribution $ p(\bfx| \bfy) $, instead of the marginal $ p(\bfx) $.
More specifically, sampling is performed by
\begin{align}
\bfx_{t+1} = \bfx_{t} + \alpha  \expect{\bfnabla_{\bfx} \log p(\bfx_{t}| \bfy)}{p(\bfy | \bfx_t)} + \bfkappa.
\label{eq:MalasDAE}
\end{align}
Fig~\ref{fig:first_fig} shows a typical example, where \method{} (top-row) successfully drives the adversarial sample to the correct cluster, while MALA (bottom-row) drives it to a wrong cluster.

To estimate the second term in Eq.\eqref{eq:MalasDAE},
we train a supervised variant of DAE, which we call a supervised DAE (sDAE),
by minimizing 
 \begin{align}
& \mathbb{E}_{p'(\bfx, \bfy)p'(\bfnu)} \Big[ \|\bfr (\bfx + \bfnu) 
 -  \bfx  \|^2 
 \notag\\
 & \qquad \qquad \qquad \qquad
- 2\sigma^2 J\left( \bfr (\bfx + \bfnu), \bfy\right)  \Big]
 .
 \label{eq:SDAEObjective}
 \end{align}
 The difference from the DAE objective \eqref{eq:DAEObjective} is in the second term,
 which is proportional to the cross entropy loss.
 With this additional term, sDAE provides the gradient estimator of the log-\emph{joint}-probability
 $\log p(\bfx, \bfy)$ averaged over the conditional training distribution.

\begin{theorem}
\label{thrm:SDAEScore}
Assume that the classifier output accurately reflects the conditional probability of the training data, i.e.,
$\widehat{\bfy}(\bfx) = p(\bfy | \bfx)$,
then the minimizer of the sDAE objective \eqref{eq:MalasDAE}
satisfies
\begin{align}
\bfr(\bfx) - \bfx  = \sigma^2 \expect{ \bfnabla_{\bfx}  \log p(\bfx, \bfy) }{p(\bfy | \bfx) }
+ O(\sigma^3).
\label{eq:ResidualScoreSupervised}
\end{align}
\end{theorem}
(Sketch of proof)
Similarly to the analysis in \cite{Alain14},
we first Taylor expand $\bfr(\bfx + \bfnu)$ around $\bfx$, and approximate
the sDAE objective \eqref{eq:SDAEObjective}
by a function similar to the objective for contractive autoencoder (CAE).
Applying the second order Euler-Lagrange equation to the approximate objective 
gives Eq.\eqref{eq:ResidualScoreSupervised} as a stationary condition.
The complete proof is given in Appendix~\ref{sec:ProofSDAEScore}.
\QED

Since $p(\bfx, \bfy) = p(\bfx| \bfy) p(\bfy)$,
if the label distribution is flat (or equivalently the number of training samples for all classes is the same), i.e., $p(\bfy) = 1 / K$,
the residual of sDAE gives an estimator for the second term in Eq.\eqref{eq:MalasDAE}:
\begin{align}
\bfr(\bfx) - \bfx  = \sigma^2 \expect{ \bfnabla_{\bfx}  \log p(\bfx| \bfy) }{p(\bfy | \bfx) }
+ O(\sigma^3).
\notag
\end{align}
The term
$\expect{ \bfnabla_{\bfx}  \log p(\bfx| \bfy) }{p(\bfy | \bfx) }$
is the gradient of the log-\emph{conditional}-distribution on the label,
where the label is estimated from the prior knowledge (the expectation is taken over the training distribution of the label, given $\bfx$).
If the number of training samples are non-uniform over the classes,
the weight (or the step size) should be adjusted so that all classes contribute equally to the training.

As discussed in Section~\ref{sec:Introduction},
\method{} falls into the category of projection methods,
in which most existing methods are outperformed by strong adversarial training methods.
Nevertheless, our experiment in Section~\ref{exp} shows its comparable (or even better under some conditions) performance than Madry \citep{madry2017towards}, the state-of-the-art adversarial training method.
We suppose that this is because of its inherent randomness---\method{} drives an input 
not always to the neighborhood of the original point (i.e., the point before the adversarial pattern was added), but any point in the cluster sharing the same label with the original image.
This large stochasticity prevents the attacker from aligning the input 
so that the \method{} \emph{stably} moves it to a targeted untrained spot where the model predicts a wrong label.

Adversarial training methods effectively fill untrained spots close to the data manifold with additional training samples, so that those weak spots of the classifier are removed.
On the other hand,
\method{} prevents samples from being classified at any untrained spot.
These two approaches are orthogonal and can naturally be combined: In the training phase, adversarial training fills most untrained spots with additional samples,
and, in the test phase, \method{} prevents the classifier from making prediction at any untrained spot
even if adversarial training failed to remove all of them.
Our experiment in Section~\ref{exp} shows that this combination outperforms all existing defense methods.


\begin{figure*}[t]
    \centering
    \begin{subfigure}[t]{0.33\textwidth}
    \centering
    \includegraphics[width=\linewidth]{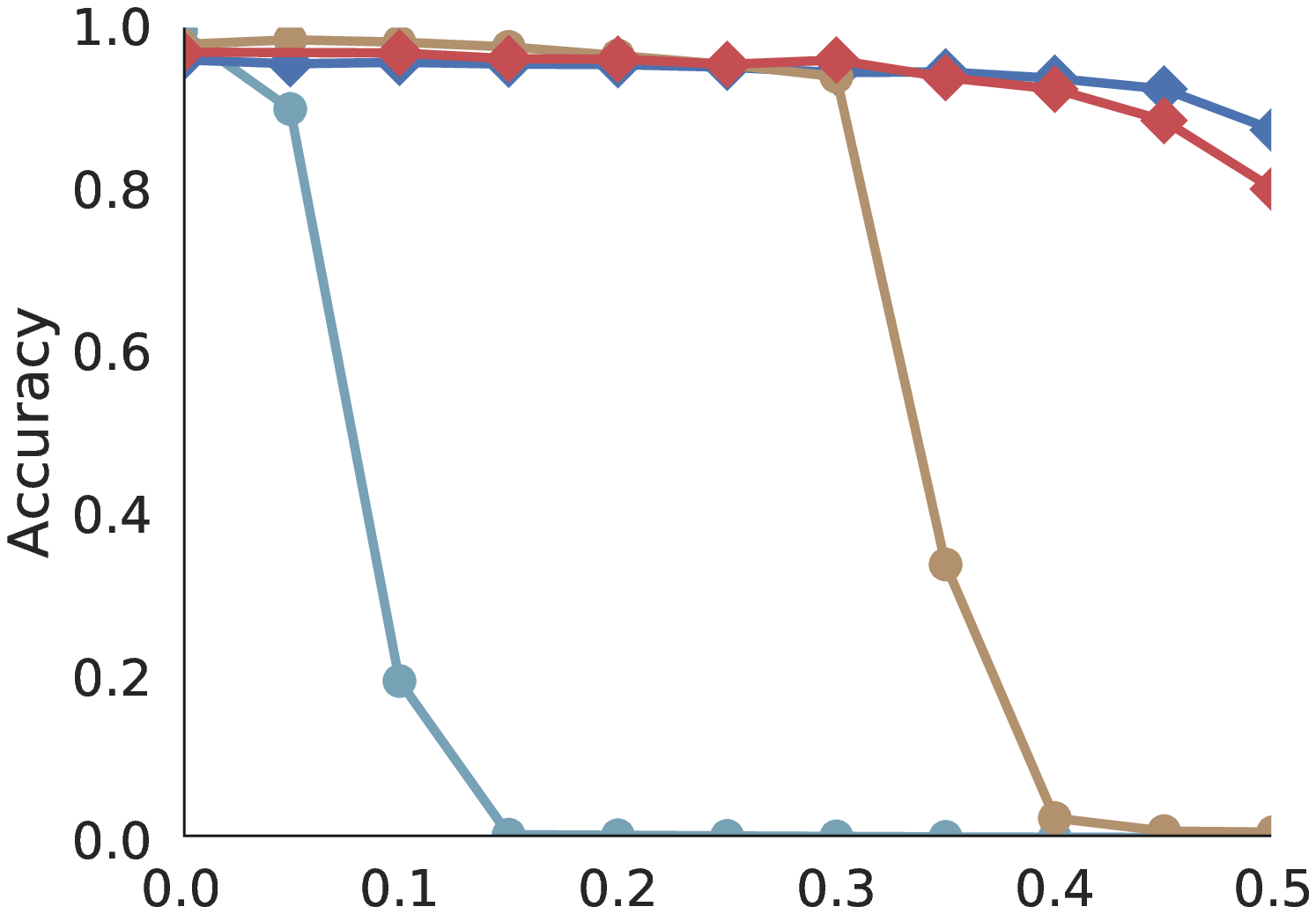}
        \caption{PGD-$L_\infty$ }
        \label{fig:pgdlinf}
    \end{subfigure}%
    ~ 
    \begin{subfigure}[t]{0.33\textwidth}
    \centering
    \includegraphics[width=\linewidth]{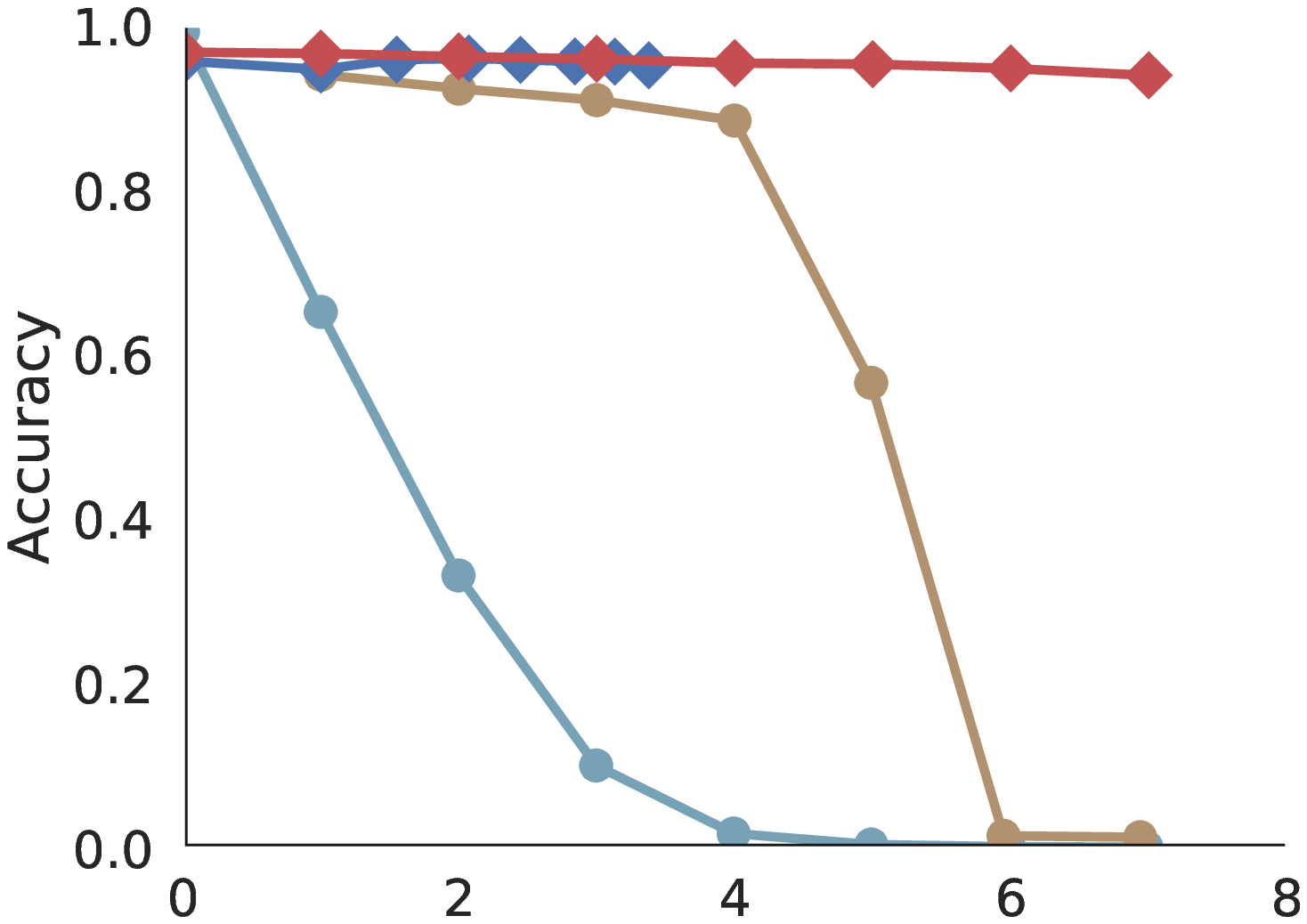}
        \caption{PGD-$L_2$}
        \label{fig:pgdl2}
    \end{subfigure}%
    ~ 
    \begin{subfigure}[t]{0.33\textwidth}
    \centering
    \includegraphics[width=\linewidth]{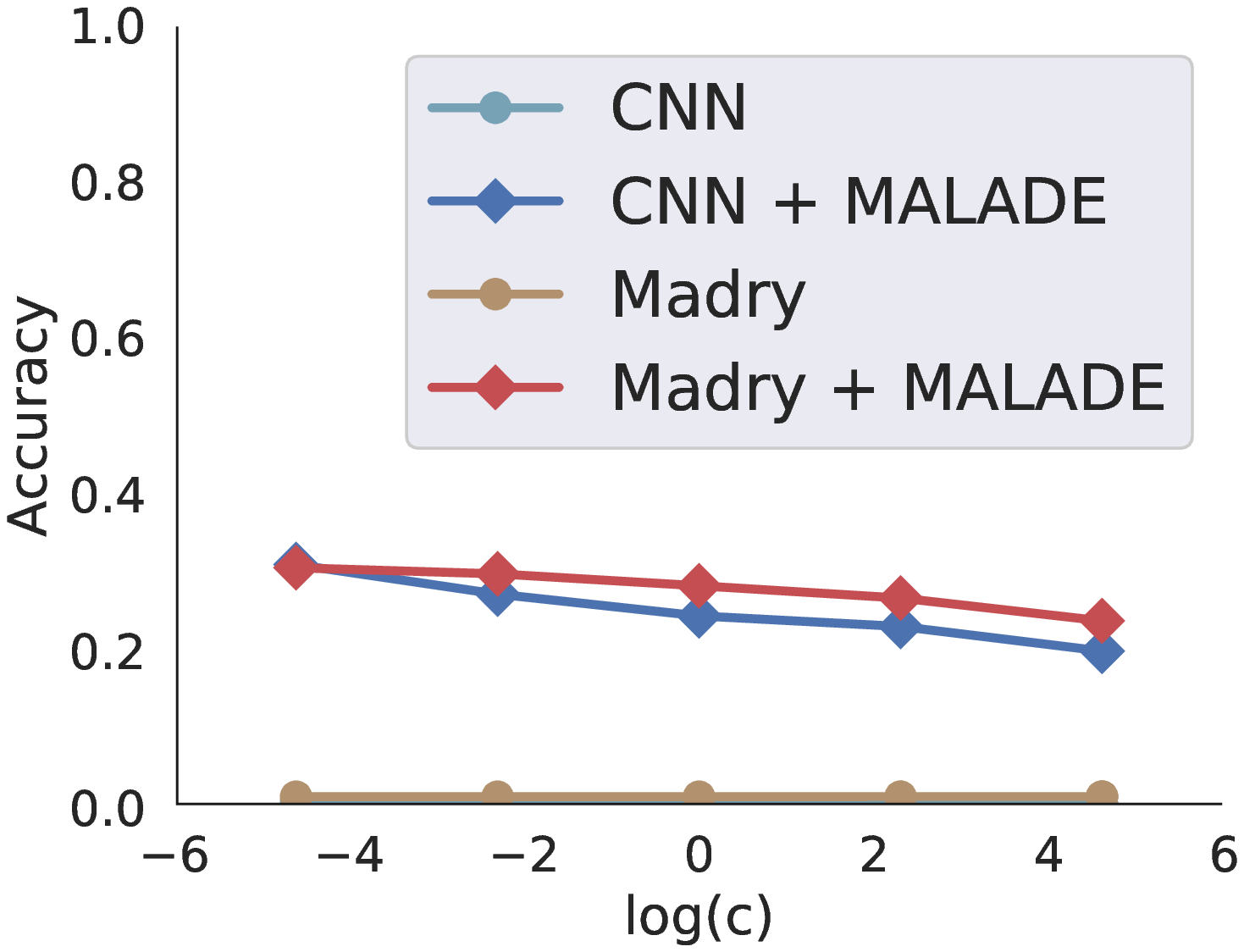}
        \caption{EAD ($(L_1+L_2)$-bounded)}
        \label{fig:ead}
    \end{subfigure}%

\caption{Classification accuracy on MNIST against (a) PGD-$L_\infty$, (b) PGD-$L_2$, and (c) EAD attacks.
The curves correspond to the original convolutional neural network (CNN) classifier, the CNN classifier protected by \method{} (CNN+MALADE), the Madry classifier \citep{madry2017towards}, and the Madry classifier protected by \method{} (Madry+MALADE).
In each plot, the horizontal axis indicates the intensity of the adversarial perturbations, i.e., $\varepsilon$ for PGD and $\log c$ for EAD.
All plots show that
\method{} significantly boosts the robustness of both classifiers.
    }
\label{fig:acc_vs_steps}
\end{figure*}

\begin{figure}[t]
    \centering
    \begin{subfigure}[t]{0.45\textwidth}
        \centering
        \includegraphics[width=\linewidth]{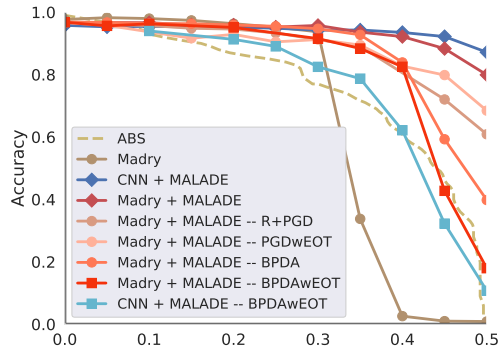}
        \caption{PGD-$L_\infty$ }
    \end{subfigure}%

\caption{
Classification accuracy on MNIST against the attacks adapted for our proposed \method{}.
The adversarial samples were created by PGD-$L_\infty$
enhanced with R regularization, BPDA, and EOT, respectively.
Although the adapted attacks reduce accuracy of \method{},
Madry+MALADE still performs better than the baselines 
(Madry and ABS).
}
\label{fig:attack_compare}


\end{figure}

\section{Experiments}
\label{exp}

In this section, we empirically evaluate 
our proposed \method{}
against various attacking strategies,
and compare it with the state-of-the-art baseline defense strategies.

\subsection{Datasets}
We conduct experiments on the following datasets:
\begin{description}
    \item [MNIST:] MNIST consists of handwritten digits from $0$-$9$. The dataset is split into training, validation and test set with $50,000$, $10,000$ and $10,000$ images, respectively. 
MNIST, in spite of being a small dataset, remains to be considered as adversarially robust. 

    \item [CIFAR10:] CIFAR10 consists of natural images for $10$ classes.  The resolution is $32\times32\times3$,
    and the dataset defines $50,000$ training images and $10,000$ test images.
    
    \item [TinyImagenet:] TinyImagenet is a subset of the large Imagenet dataset consisting of 200 classes out of the 1000 classes. Each class contains 500 training images and 50 test images. The images are subsampled to have a resolution of $64\times64\times3$.
\end{description}

\subsection{Attacking Strategies}
We evaluate defense methods against the following attacking strategies:
\begin{description}
\item [Whitebox Attacks:]
We chose PGD \citep{moosavi2016deepfool}, CW \citep{ilyas2017query}, MIM \citep{brendel2017decision}, 
and EAD \citep{zhao2018admm} as state-of-the-art whitebox attacks. 
Indeed they have broken down many recent defense strategies \citep{athalye2018obfuscated}, and MIM is the winner of NIPS 2017 competition on adversarial attacks \citep{dong2018mim}.

\item [Blackbox Attacks:]
We chose Boundary Attack \cite{brendel2017decision} and Salt and Pepper Attack \cite{hendrycks2018benchmarking}
as state-of-the-art blackbox attacks.
We omit distillation attacks \citep{Papernot, Papernotb} because they are in principle weaker than the corresponding whitebox attacks.
We add Transfer Attack \cite{tramer2017space} under the blackbox scenario,
where the attackers craft the data on different defense strategies.


\end{description}


In our experiment, 
we pay careful attention to \emph{fairness} of evaluation.
Recent papers \citep{engstrom2018evaluating, athalye2018obfuscated, uesato2018adversarial, carlini2019evaluating} reported
on inappropriate evaluation of defense strategies in literature.
Most of the cases, attacking samples were created
with inappropriate parameter setting, e.g., the maximum number of iterations is too small to converge,
or attacking methods were not properly \emph{adapted}
so that the they are effective to the proposed defense method.
To avoid this, we first make sure that the attacking methods are applied with appropriate parameter setting.  Specifically, we set the maximum number of iterations to a sufficient large value, and confirm the convergence
(see Appendix~\ref{sec:App.ParameterSetting}).
Furthermore, we explore possible attacking methods suitable against our proposed \method{},
by combining following ideas with the general whitebox strategies:
\begin{description}
\item [Whitebox Attacks adapted to \method{}:]
We consider R regularization \citep{frosst2018darccc},
BPDA \citep{athalye2018obfuscated}, and EOT \citep{athalye2018obfuscated}
to be suitable strategies for breaking down \method{}, 
which is equipped with stochastic denoising (projection) step.
They are used in combination with general whitebox attacks, e.g., R+PGD, PGD with EOT (PGDwEOT), etc.
\end{description}

\subsection{Baseline Defense Strategies}

We choose two adversarial training methods, 
Madry \citep{madry2017towards} and ALP \cite{kannan2018adversarial},
as the state-of-the-art baseline defense methods to be compared with
our proposed \method{}.
Due to the availability of pretrained networks,
we evaluate Madry on MNIST and CIFAR10,%
\footnote{\url{https://github.com/tensorflow/models/tree/master/research/adversarial\_logit\_pairing}}
and ALP on TinyImagenet.%
\footnote{\url{https://github.com/MadryLab/mnist\_challenge}}

\subsection{Results on MINST}


We first show our extensive experiments on MNIST.
Fig.~\ref{fig:acc_vs_steps} shows classification accuracy of the original convolutional neural network (CNN) classifier,
the CNN classifier protected by \method{} (CNN+MALADE),
the Madry classifier, 
i.e., the classifier trained with adversarial samples,
and the Madry classifier protected by \method{} (Madry+MALADE).
The adversarial samples were created by (a) PGD-$L_\infty$, (b) PGD-$L_2$,
and (c) EAD attacks,
where PGD-$L_p$ denotes the $L_p$-bounded PGD attack.
Note that EAD can be seen as an $(L_1+L_2)$-bounded attack,
producing sparse adversarial patterns.
In each plot in Fig.~\ref{fig:acc_vs_steps}, the horizontal axis indicates to the intensity of the adversarial perturbations, i.e., $\varepsilon$ for PGD and $\log c$ for EAD.
The {Madry} classifier was trained on the adversarial samples created by PGD-$L_\infty$ for $\varepsilon = 0.3$.

As expected, Madry is robust against PGD-$L_\infty$ up to $\varepsilon \leq0.3$. 
Consistently with the author's report \citep{madry2017towards},
it is also robust against PGD-$L_2$ up to 
$\varepsilon\leq 4.5$. 
However, Madry is broken down by PGD-$L_\infty$ for $\varepsilon>0.4$ and PGD-$L_2$ for $\varepsilon>6$. 
Furthermore, EAD, which was proposed more recently than Madry, completely 
 breaks down Madry.
On the other hand,
our proposed \method{} is robust against PGD-$L_\infty$ and PGD-$L_2$ 
in a wide range of $\varepsilon$, 
and is not completely broken down by EAD.


\begin{table*}[t]
  \caption{Summary of classification performance on MNIST. 
  For each attacking scenario, i.e., whitebox/blackbox and bounds for the amplitude of adversarial patterns, the lowest row gives the \emph{worst case} result over the considered attacking strategies.
  }
  \label{table:mnist}
  \centering
  \resizebox{\columnwidth}{!}{%
  
  \begin{tabular}{l c c c c c c c  }
    \toprule
     Setting
     & Condition
     & Attack 
     & CNN
     & Madry
     & CNN + \method{} & 
     Madry + \method{} \\
    
    \midrule
    
            &     & FGSM 						  & 11.77   & 97.52 & 93.54 & 95.59\\
&       & PGD 							  & 0.00    & 93.71 & 94.22 & 95.76\\   
&      $L_\infty$           & R+PGD						  & -           & -     & 92.65     & 93.51\\   
&     $\varepsilon=0.3$            & BPDA						      & -           & -     & 84.74     & 94.70\\   
&                 & BPDAwEOT					      & -           & -     & 82.48     & 91.54\\   
&                 & MIM 	    				 	  & 0.00        & 97.66 & 94.32 & 94.53\\   
        \cmidrule(r){3-7}
&                &  \emph{worst case}                               & 0.00 & \textbf{93.71} &  82.48 &  91.54 \\
        \cmidrule(r){2-7}
&                & PGD 							  & 0.00    & 0.02 & 93.51 & 92.18\\   
\emph{whitebox}& $L_\infty$      & BPDA						      & -           & -     & 66.16     & 83.90\\   
& $\varepsilon=0.4$   & BPDAwEOT					      & -           & -     & 62.15     & 80.65\\   
        \cmidrule(r){3-7}
&                &   \emph{worst case}                  & 0.00 & 0.02 &  62.15 &  \textbf{80.65} \\
        \cmidrule(r{2em}){2-7} 
&                 & FGM 							  &  30.79 & 97.68 & 94.68 & 96.05\\
&     $L_2$       & PGD 							  &  0.01 & 92.68 & 95.91 & 96.76 \\   
&    $\varepsilon=4$            & CW      						  &  0.00 & 85.53 & 90.07 & 91.14\\   
        \cmidrule(r){3-7}
&                &   \emph{worst case}                               & 0.0& 85.53 &  90.07 &  \textbf{91.14} \\
                 
        \cmidrule(r){2-7}
& $L_2$-$L_1$     & EAD       				      & 0.00 & 0.01 & \textbf{31.00} & 30.59 \\      
& $\beta=0.01$ and $c=0.01$     &       				      &  &  &  &  \\      

       
        \midrule
  & $\rho=0.25$ & SaltnPepper & 36.49 & 41.61 & 80.41 &  {80.72} \\
 \emph{blackbox}& $T=5,000$ & Boundary Attack & 32.39 & 1.10 & 93.79 & {95.80} \\
        \cmidrule(r){3-7}
&                &   \emph{worst case}                               & 32.39& 1.10 &  80.41 &  \textbf{80.72} \\


    \bottomrule

    \end{tabular}
  \vspace{1mm}
 }
 \end{table*}

Next, we investigate the robustness against attacking strategies,
which we elaborated for attacking our own method.
Fig.~\ref{fig:attack_compare} shows the performance of Madry+MALADE against PGD-$L_\infty$ enhanced by R regularization, BPDA, and EOT, respectively,
which are considered to be effective against \method{}.
For comparison, we show the performance of Madry and ABS \citep{schott2018robust}
against the plain PGD-$L_\infty$.
We see that, although the elaborated attacks reduce the accuracy of Madry+MALADE to some extent, their effect is limited, and even in the worst case, i.e., against BPDA with EOT (BPDAwEOT), 
Madry+MALADE performs better than the baseline methods, i.e., Madry and ABS.
The effect of EOT is visualized in Fig.\ref{fig:grad_obfusc_mnist} in Appendix.

Table~\ref{table:mnist} summarizes classification performance of defense strategies against various attacks.
For each attacking scenario, i.e., whitebox/blackbox and bounds for the amplitude of adversarial patterns, the lowest row gives the \emph{worst case} result over the considered attacking strategies,
which should be seen as a robustness criterion of defense strategy---a good defense method must be robust against any attacking strategy.

Madry performs best under the whitebox scenario with $L_\infty$-bounded adversary for $\varepsilon = 0.3$---the setting against which Madry was trained to be robust.
However, Madry+MALADE shows comparable performance under this setting.
For the other cases including larger $\varepsilon$ and different norm bounds, Madry+MALADE, as well as CNN+MALADE, significantly outperforms Madry.




Table~\ref{table:mnist} also shows results under the blackbox scenario
with the state-of-the-art attacking strategies, 
SaltnPepper and Boundary attack.
The table clearly shows high robustness of \method{} against those attacks,
while Madry exhibits its vulnerability against them.
 

\begin{table} [t]
\caption{Classification performance on MNIST under the transfer attack with $L_\infty$-bounded adversary for $\varepsilon=0.4$. 
The columns correspond to the classifiers of which the robustness is measured,
while the rows corresponds to the classifiers against which the adversarial
samples are crafted.
The bottom row shows the \emph{worst case} result for defender,
i.e., the most successful attack by an attacker who
has full knowledge on the three other classifiers other than the target classifier.
}
\label{table:transfer_attack}
\centering
  
\begin{tabular}{|c|c|c|c|c|}
\hline
 & CNN & Madry & CNN +  & Madry + \\
 &  &  & \method{} & \method{}\\
 
\hline
CNN & - & 81.20 & 90.11 & 92.84\\
\hline
Madry & 26.92 & - & 76.87 & 73.69\\
\hline
CNN +  &  &  &  & \\
\method{} & 9.19 & 77.55 & - & 71.40\\
\hline
Madry +   &  &  &  & \\
\method{} & 26.12 & 63.47 & 78.20 & -\\
\hline
\emph{worst} &  &  &  & \\
\emph{case} & 9.19 & 63.47 & \textbf{76.87} & {71.40}\\
\hline
\end{tabular}
\label{tbl:Transfer}
\end{table}

Table~\ref{tbl:Transfer} evaluates robustness of the four classifiers, CNN, Madry, CNN+MALADE, and Madry+MALADE, against transfer attacks.
Assume for example that, when CNN is attacked, the attacker has full knowledge
on Madry, CNN+MALADE, and Madry+MALADE but not on CNN.%
\footnote{This includes the case that the attacker simply does not know that the target classifier is the plain CNN.
}
In such a case, a reasonable strategy for the attacker is to craft adversarial samples
against each of the known classifiers, i.e., Madry, CNN+MALADE, and Madry+MALADE, and use them to attack the target classifier, i.e., CNN.
This particular example corresponds to the second column of Table~\ref{tbl:Transfer},
where the bottom row shows the \emph{worst case} result for the defender, i.e., 
the most successful attack.
For the attacking method, the strongest one in Table~\ref{table:mnist} was chosen,
i.e., PGD when adversarial samples are crafted against CNN and Madry,
and BPDAwEOT when adversarial samples are crafted against CNN+MALADE and Madry+MALADE.
Also in this transfer attack scenario,
we see that our \method{} outperforms the state-of-the-art Madry classifier.



\begin{figure}[h]
    \centering
    \begin{subfigure}[t]{0.45\textwidth}
    \centering
    \includegraphics[width=\linewidth]{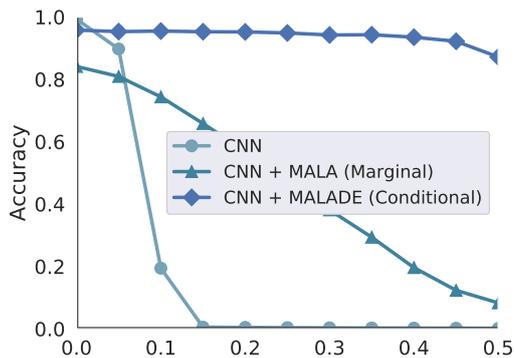}
        \caption{PGD-$L_\infty$ }
    \end{subfigure}%
    
\caption{
Comparison between \method{} and MALA (with marginal distribution).
The superiority of  \method{} to MALA (maginal) 
is clearly seen,
which 
proves the benefit of using perceptual boundary information
in driving adversarial samples into high density areas.
} 
\label{fig:mala}
\end{figure}

\begin{table*}[t]
  \caption{ Summary of classification performance on CIFAR10. 
  For PGD attack, the number of steps was fixed at $100$. 
  For blackbox attacking scenario, 
  the lowest row gives the \emph{worst case} result over the considered attacking strategies.}
  
  \label{table:cifar}
  \centering
   \resizebox{\columnwidth}{!}{%
  
  \begin{tabular}{l c c c c c c  c}
    \toprule
    Setting
    & Condition
    & Attack 
     & CNN
     & Madry
     & CNN + \method{}
     & Madry + \method{} \\
     
    \midrule

       & $L_\infty$ $\varepsilon=8$  & PGD  & 0.00 & 32.86   & 0.48   &  \textbf{33.89} \\   
       \emph{whitebox}  &  $L_\infty$ $\varepsilon=16$ & PGD & 0.00 & 10.28  & 0.16  &  \textbf{11.46}  \\  
    \midrule
           & $\rho=0.05$ & SaltnPepper      & 13.54 & 11.09 & 15.19  & 11.12 \\
         \emph{blackbox} & $T=10,000$ & Boundary Attack & 12.76 & 34.97 & 17.72 & 73.57 \\
        \cmidrule(r){3-7}
           &  & \emph{worst case}  & 12.76 & 11.09   & \textbf{15.19}   & 11.12 \\   
    
    \bottomrule

    \end{tabular}
  \vspace{1mm}
}
\end{table*}

\begin{table*}[t]
\caption{ Summary of classification performance on TinyImagenet. 
  For PGD attack, the number of steps was fixed at $100$. 
  For blackbox attacking scenario, 
  the lowest row gives the \emph{worst case} result over the considered attacking strategies.}
  
  
  \label{table:tinyimagenet}
  \centering
  \resizebox{\columnwidth}{!}{%
  
  \begin{tabular}{l c c c c c c c }
    \toprule
  
    Setting & Condition & Attack & CNN & ALP & CNN + \method{} & ALP + \method{} \\
     
    \midrule
    
      
      & $L_\infty$ $\varepsilon=1$ & PGD   & 9.49 & 27.05 & 10.07 & \textbf{31.80} \\ 
     \emph{whitebox} & $L_\infty$ $\varepsilon=4$& PGD & 0.20  & 12.78 & 0.26 & \textbf{13.23} \\   
       \midrule
          & $\rho=0.05$  & SaltnPepper        & 2.45 & 9.91 & 6.23  & 9.94 \\
        \emph{blackbox} & $T=10,000$ & Boundary Attack & 9.12  & 7.38 & 63.54 & 44.35 \\
        \cmidrule(r){3-7}
          &  & \emph{worst case}  & 2.45 & 7.38 & 6.23   & \textbf{9.94}  \\   
    \bottomrule

    \end{tabular}
  \vspace{1mm}
}
\end{table*}

Summarizing our experimental results on MNIST,
our proposed \method{} showed excellent performance under different attacking scenarios,
and different threat models (bounds for adversarial patterns).
Adversarial examples crafted by various strategies can be found 
in Appendix~\ref{images_mnist}.

Before moving to experiments on larger scale data,
we show the importance of our new development, i.e., sDAE providing
the gradient for the conditional distribution $p(\bfx | \bfy)$.
To this end, we compare in Fig.~\ref{fig:mala} the classification accuracy of MALADE and MALA (with marginal distribution $p(\bfx)$) against PGD-$L_\infty$.
MALA (marginal) shows robustness against adversarial attack with high intensity
(compared to the original CNN classifier).
However, the lack of perceptual boundary information clearly degrades the performance,
and thus MALA (marginal) is significantly outperformed by \method{} over all intensity range.
This supports our initial hypothesis explained in Fig.\ref{fig:cartoon_fig}---%
MALA drives a sample to a neighboring cluster which does not necessarily share the correct label---%
and imply that
cases similar to the example shown in Fig.\ref{fig:first_fig}
often happen.

\subsection{Results on CIFAR10 and TinyImagenet}

Here we apply our defense strategy for larger datasets.
Tables~\ref{table:cifar} and \ref{table:tinyimagenet}
report on classification accuracy on CIFAR10 and on TinyImagenet, respectively,
against whitebox (PGD-$L_\infty$) and blackbox (SaltnPepper and Boundary) attacks. 
On TinyImagenet, we show the performance of ALP \citep{kannan2018adversarial}
as the baseline defense method.
Similarly to Madry+MALADE, it is straightforward to combine \method{} and ALP to form ALP+MALADE.



Overall, classification accuracy is not high, which reflects the fact that the state-of-the-art has not achieved satisfactory level of robustness against adversarial attacks in the scale of CIFAR10 and TinyImagenet datasets.
However, the observation that 
\method{} consistently improves the robustness of classifiers
implies that our approach is a hopeful direction for solving the issue of defense against adversarial attacks in large scale problems.

\section{Conclusion}
\label{conclusion}

Adversarial attacks against deep learning models change a sample so that human perception does not allow to detect the change. However, the classifier is compromised and yields a false prediction. 
Many defense strategies attempted to alleviate this effect by pre-processing,
projection,
adversarial training,
or robust optimization. 
However most of the defenses have been broken down by newer attacks or variations of  existing attacks. 

In this work, we have proposed to use the Metropolis-adjusted Langevin algorithm (MALA) which is guided through a supervised DAE---MALA for DEfense (\method{}). 
This framework allows us to drive adversarial samples towards the underlying data manifold and thus towards the high density regions of the data generating distribution,
where the nonlinear learning machine is supposed to be trained well with sufficiently many training data. 
In this process, the gradient is computed not based on the \emph{marginal} input distribution
but on the \emph{conditional} input distribution given output, 
and it is estimated by a novel supervised DAE.
This prevents \method{} from driving samples into a neighboring cluster with a wrong label,
and gives rise to high generalization performance that significantly reduces the effect of adversarial attacks. 
We have empirically showed that \method{} is fairly robust---%
it compares favorably or significantly outperforms the state-of-the-art defense methods \citep{madry2017towards,kannan2018adversarial}
under different adversarial scenarios and attacking strategies.

Supposing that the difficulty lies in the complex structure of untrained spots close to the data manifold,
which increases along with the increase in data size and complexity of the model,
we see \method{}
as a hopeful defence strategy, which effectively removes a lot of untrained spots simultaneously without using additional training samples to fill those spots one by one.
However, 
further efforts must be made, e.g., 
in stabilizing the prediction by majority voting
from a collection of the generated samples after burn-in,
and in developing tools which robustly estimate the gradient
in high dimensional space,
which are left as future work.
Other future work includes
analyzing the attacks and defenses using interpretation methods \citep{MonDSP18, lapuschkin2019unmasking}, 
and applying the supervised DAE to other applications such as federated or distributed learning \citep{mcmahan2016communication, SatArXiv18}.

\subsubsection*{Acknowledgments}
This work was supported by the Fraunhofer Society under the MPI-FhG collaboration project ``Theory \& Practice for Reduced Learning Machines''.
This work was also supported by the German Ministry for Education and Research (BMBF) as Berlin Big Data Center (01IS18025A) and Berlin Center for Machine Learning (01IS18037I), 
the German Research Foundation (DFG) as Math+: Berlin Mathematics Research Center (EXC 2046/1, project-ID: 390685689),
and the Information \& Communications Technology Planning \& Evaluation (IITP) grant funded by the Korea government (No. 2017-0-00451, No. 2017-0-01779).


\bibliographystyle{unsrt}

\bibliography{nips_2018}





\newpage


\onecolumn
\appendix

\section{Proof of Theorem~\ref{thrm:SDAEScore}}
\label{sec:ProofSDAEScore}

sDAE is trained so that the following functional is minimized with respect to the function 
$\bfr:\mathbb{R}^L \mapsto \mathbb{R}^L$:
\begin{align}
\Obj(\bfr) &=\expect{ \|\bfr (\bfx + \bfnu) -  \bfx  \|^2 - 2\sigma^2 J\left(\bfr (\bfx + \bfnu), \bfy \right)  }
{p'(\bfx, \bfy)p'(\bfnu)},
\label{eq:SDAEObjectiveAp}
\end{align}
which is a finite sample approximation to the true objective
\begin{align}
\Obj(\bfr) &= \int  \left( \| \bfr (\bfx + \bfnu)- \bfx  \|^2  - 2 \sigma^2  \expect{\log p\left(\bfy |\bfr (\bfx+ \bfnu)\right)}{p(\bfy | \bfx)}      \right)p(\bfx)    \mathcal{N}_L (\bfnu; \bfzero, \sigma^2 \bfI) d\bfx  d\bfnu.
\label{eq:DAEObjectiveTrue}
\end{align}

We assume that $\bfr(\bfx)$ and $ p\left(\bfy |\bfx\right) $  are analytic functions with respect to $\bfx$.
For small $\sigma^2$, 
the Taylor expansion of the $l$-th component of $\bfr$ around $\bfx$ gives 
\begin{align}
r_l (\bfx + \bfnu)
&= r_l (\bfx) + \bfnu^\T \frac{\partial r_l}{ \partial \bfx}  + \frac{1}{2} \bfnu^\T \frac{\partial^2 r_l}{ \partial \bfx \partial \bfx} \bfnu + O(\sigma^3),
\notag
\end{align}
where $ \frac{\partial^2 f}{ \partial \bfx \partial \bfx} $ is the Hessian of a function $f(\bfx)$.
Substituting this into Eq.\eqref{eq:DAEObjectiveTrue}, we have
\begin{align}
\Obj 
&=  \int   \Bigg\{  \sum_{l=1}^L \left( r_l (\bfx) + \bfnu^\T \frac{\partial r_l}{ \partial \bfx}  + \frac{1}{2} \bfnu^\T \frac{\partial^2 r_l}{ \partial \bfx \partial \bfx} \bfnu - x_l \right)^2
 \notag\\
& \qquad \qquad \qquad \qquad \qquad
 - 2 \sigma^2 \expect{\log p\left(\bfy |\bfr (\bfx)\right)  }{p(\bfy | \bfx)} 
\Bigg\}
 p(\bfx)   d\bfx  \mathcal{N}_L (\bfnu; \bfzero, \sigma^2 \bfI) d\bfnu
+ O(\sigma^3)
\notag\\
&=  \int \Bigg\{ \sum_{l=1}^L \left(\left(   r_l (\bfx)- x_l\right)^2  + \left(  r_l (\bfx)- x_l \right) \bfnu^\T \frac{\partial^2 r_l}{ \partial \bfx \partial \bfx} \bfnu 
+ \frac{\partial r_l}{ \partial \bfx}^\T \bfnu  \bfnu^\T \frac{\partial r_l}{ \partial \bfx} 
\right)
 \notag\\
& \qquad \qquad \qquad \qquad \qquad
 - 2 \sigma^2 \expect{\log p\left(\bfy |\bfr (\bfx)\right)  }{p(\bfy | \bfx)}  
\Bigg\}
 p(\bfx)   d\bfx  \mathcal{N}_L (\bfnu; \bfzero, \sigma^2 \bfI) d\bfnu
+ O(\sigma^3)
\notag\\
&=  \int \Bigg\{\sum_{l=1}^L \left( \left(   r_l (\bfx) - x_l\right)^2  + \sigma^2  \left(r_l (\bfx) -  x_l \right) \tr \left( \frac{\partial^2 r_l}{ \partial \bfx \partial \bfx} \right) 
+\sigma^2 \left\| \frac{\partial r_l}{ \partial \bfx}  \right\|^2
\right)
 \notag\\
& \qquad \qquad \qquad \qquad \qquad
 - 2 \sigma^2 \expect{\log p\left(\bfy |\bfr (\bfx)\right)  }{p(\bfy | \bfx)} 
\Bigg\}
 p(\bfx)   d\bfx 
+ O(\sigma^3).
\label{eq:ObjectiveAsym}
\end{align}

Thus, the objective functional \eqref{eq:DAEObjectiveTrue} can be written as
\begin{align}
\Obj(\bfr)
&=  \int \ObjInt  d\bfx + O(\sigma^3),
\label{eq:ObjectiveAnal}
\end{align}
where
\begin{align}
\ObjInt
&= \Bigg\{\sum_{l=1}^L\left( \left(  r_l (\bfx)  - x_l+ \sigma^2 \tr \left( \frac{\partial^2 r_l}{ \partial \bfx \partial \bfx} \right)\right)  \left( r_l (\bfx) - x_l \right)
+\sigma^2 \left\| \frac{\partial r_l}{ \partial \bfx}  \right\|^2
\right)
 \notag\\
& \qquad \qquad \qquad \qquad \qquad
-2 \sigma^2 \expect{\log p\left(\bfy |\bfr(\bfx)\right)  }{p(\bfy | \bfx)} 
\Bigg\}
 p(\bfx) .
\label{eq:ObjectiveAnalL}
\end{align}

We can find the optimal function minimizing the functional \eqref{eq:ObjectiveAnal} by using \emph{calculus of variations}.
The optimal function satisfies the following Euler-Lagrange equation: for each $l = 1, \ldots, L$,
\begin{align}
\frac{\partial \ObjInt}{ \partial r_l} 
- \sum_{m=1}^L \frac{\partial }{ \partial x_m} \frac{\partial \ObjInt}{ \partial (\bfr_l')_m} 
+  \sum_{m=1}^L   \sum_{m'=m+1}^L  \frac{\partial^2 }{  \partial x_m \partial x_{m'}} \frac{\partial \ObjInt}{ \partial (\bfR_l'')_{m, m'}}
=0,
\label{eq:EulerLagrange}
\end{align}
where $\bfr_l' = \frac{\partial r_l}{ \partial \bfx} \in \mathbb{R}^L$ is the gradient (of $r_l$ with respect to $\bfx$) and 
$\bfR_l'' = \frac{\partial r_l}{  \partial \bfx \partial \bfx} \in \mathbb{R}^{L \times L}$ is the Hessian.

We have
\begin{align}
\frac{\partial \ObjInt}{\partial r_l} 
&=  \left\{ 2 \left(   r_l (\bfx) - x_l\right)   + \sigma^2 \tr \left( \frac{\partial^2 r_l}{ \partial \bfx \partial \bfx} \right) 
- 2 \sigma^2\frac{ \partial \expect{\log p\left(\bfy |\bfr(\bfx)\right)  }{p(\bfy | \bfx)} }{\partial r_l} 
\right\}
 p(\bfx),
\notag\\
\frac{\partial \ObjInt}{\partial (\bfr'_l)_m} 
&= 2 \sigma^2  \frac{\partial r_l}{ \partial x_m}  p(\bfx) ,
\notag\\
 \frac{\partial \ObjInt}{ \partial (\bfR_l'')_{m, m'}}
&= \delta_{m, m'}  \sigma^2  \left( r_l (\bfx) - x_l \right)   p(\bfx),
\notag
\end{align}
and therefore
\begin{align}
\frac{\partial}{\partial x_m}
\frac{\partial \ObjInt}{\partial (\bfr'_l)_m} 
&=2 \sigma^2 \left(   \frac{\partial^2 r_l}{ \partial x_m^2} p(\bfx) +     \frac{\partial r_l}{ \partial x_m} \frac{\partial p(\bfx)}{\partial x_m} \right),
\notag\\
\frac{\partial^2}{\partial x_m \partial x_{m'}}
 \frac{\partial \ObjInt}{ \partial (\bfR_l'')_{m, m'}}
&=  \sigma^2\delta_{m, m'} 
 \frac{\partial}{ \partial x_{m}} 
\left(  \left(  \frac{\partial r_l}{ \partial x_{m'}} - \delta_{l, m'}\right)  p(\bfx) +   \left(  r_l (\bfx) - x_l \right) \frac{\partial p(\bfx)}{\partial x_{m'}} \right)
\notag\\
& \hspace{-10mm}=  \sigma^2\delta_{m, m'} 
\left(  \frac{\partial^2 r_l}{ \partial x_m^2}   p(\bfx) 
+ 2 \left( \frac{\partial r_l}{ \partial x_m} - \delta_{l, m} \right) \frac{\partial p(\bfx)}{ \partial x_{m}} 
 +   \left(   r_l (\bfx) - x_l\right) \frac{\partial^2 p(\bfx)}{\partial x_m^2}
  \right),
 \notag
\end{align}
where $ \delta_{m, m'}$ is the Kronecker delta.
Substituting the above into Eq.\eqref{eq:EulerLagrange}, we have
\begin{align}
& \textstyle \left\{ 2 \left(  r_l (\bfx) - x_l\right)   + \sigma^2 \tr \left( \frac{\partial^2 r_l}{ \partial \bfx \partial \bfx} \right) - 2 \sigma^2\frac{ \partial \expect{\log p\left(\bfy |\bfr(\bfx)\right)  }{p(\bfy | \bfx)} }{\partial r_l}  \right\} p(\bfx)
\notag\\
& \quad
\textstyle
+ \sigma^2 \sum_{m=1}^L   \sum_{m'=m+1}^L  
\delta_{m, m'} 
\left(  \frac{\partial^2 r_l}{ \partial x_m^2}   p(\bfx) 
+ 2 \left( \frac{\partial r_l}{ \partial x_m} - \delta_{l, m}\right)  \frac{\partial p(\bfx)}{ \partial x_{m}} 
 +   \left(  r_l (\bfx) - x_l \right) \frac{\partial^2 p(\bfx)}{\partial x_m^2}
  \right)
  \notag \\
& \qquad  \qquad \qquad  \qquad 
\textstyle
-2 \sigma^2 \sum_{m=1}^L 
 \left(  \frac{\partial^2 r_l}{ \partial x_m^2} p(\bfx) +     \frac{\partial r_l}{ \partial x_m} \frac{\partial p(\bfx)}{\partial x_m} \right)
=0,
 \end{align}
 and therefore
 \begin{align}
&  \left( r_l (\bfx) - x_l - \sigma^2 \frac{ \partial \expect{\log p\left(\bfy |\bfr(\bfx)\right)  }{p(\bfy | \bfx)} }{\partial r_l} \right)
 \left(  1+ 
   \frac{ \sigma^2 }{2 p(\bfx) }
\tr \left(
\frac{\partial^2 p(\bfx)}{\partial \bfx \partial \bfx}
  \right) \right)
-\sigma^2\frac{1}{p(\bfx) } \frac{\partial p(\bfx)}{ \partial x_{l}} 
=0.
\notag
\end{align}

Since
\begin{align}
\frac{\partial \log p(\bfx)}{\partial x_m } 
&=
\frac{1}{p(\bfx)}
\frac{\partial p(\bfx)}{\partial x_m } ,
\notag\\
\frac{\partial^2 \log p(\bfx)}{\partial x_m \partial x_{m'}} 
&=
\frac{\partial}{\partial x_{m'}} 
\left(
\frac{1}{p(\bfx)}
\frac{\partial p(\bfx)}{\partial x_m } 
\right)
\notag\\
&=
-
\frac{\partial \log p(\bfx)}{\partial x_{m'} } 
\frac{\partial \log  p(\bfx)}{\partial x_m } 
+
\frac{1}{p(\bfx)}
\frac{\partial^2 p(\bfx)}{\partial x_m \partial x_{m'} } ,
\notag \\
\frac{ \partial \expect{\log p\left(\bfy |\bfr(\bfx)\right)  }{p(\bfy | \bfx)} }{\partial r_l}
&= \expect{\frac{ \partial \log p\left(\bfy |\bfx\right)  }{\partial x_l}  \bigg|_{\bfx = \bfr(\bfx)}}{p(\bfy | \bfx) },
\notag
\end{align}
we have
\begin{align}
& \textstyle \left( r_l (\bfx) - x_l - \sigma^2\expect{\frac{ \partial \log p\left(\bfy |\bfx\right)  }{\partial x_l} \Big|_{\bfx = \bfr(\bfx)}}{p(\bfy | \bfx) } \right)
 \left(  1+ 
   \frac{ \sigma^2 }{2 }
   \left(
\tr \left(
\frac{\partial^2 \log p(\bfx)}{\partial \bfx \partial \bfx}
  \right)
  + \left\|   \frac{\partial \log p(\bfx)}{\partial \bfx }  \right\|^2
 \right)  \right)
 \notag\\
 & \qquad \qquad\qquad\qquad \qquad
 \textstyle
-\sigma^2  \frac{\partial \log p(\bfx)}{ \partial x_{l}} 
=0,
\notag
\end{align}
and therefore
\begin{align}
r_l (\bfx) - x_l 
&=
\sigma^2 \expect{\frac{ \partial \log p\left(\bfy |\bfx\right)  }{\partial x_l} \bigg|_{\bfx = \bfr(\bfx)}}{p(\bfy | \bfx) }
\notag\\
& \qquad \qquad
+
\sigma^2  \frac{\partial \log p(\bfx)}{ \partial x_{l}} 
 \left(  1+ 
   \frac{ \sigma^2 }{2 }
   \left(
\tr \left(
\frac{\partial^2 \log p(\bfx)}{\partial \bfx \partial \bfx}
  \right)
  + \left\|   \frac{\partial \log p(\bfx)}{\partial \bfx }  \right\|^2
 \right)  \right)^{-1}
 \notag\\
 &=
 \sigma^2 \expect{\frac{ \partial \log p\left(\bfy |\bfx\right)  }{\partial x_l} \bigg|_{\bfx = \bfr(\bfx)}}{p(\bfy | \bfx) }
 +
\sigma^2  \frac{\partial \log p(\bfx)}{ \partial x_{l}} 
+ O(\sigma^4).
 \notag
 \end{align}
 Taking the asymptotic term in Eq.\eqref{eq:ObjectiveAnal} into account,
we have
 \begin{align}
r_l (\bfx) - x_l 
 &=
 \sigma^2 \expect{\frac{ \partial \log p\left(\bfy |\bfx\right)  }{\partial x_l} \bigg|_{\bfx = \bfr(\bfx)}}{p(\bfy | \bfx) }
 +
\sigma^2  \frac{\partial \log p(\bfx)}{ \partial x_{l}} 
+ O(\sigma^3),
 \notag
 \end{align}
which implies that $\bfr(\bfx) = \bfx + O(\sigma^2) $.
Thus, we conclude that 
 \begin{align}
r_l (\bfx) - x_l 
 &=
 \sigma^2 \expect{\frac{ \partial \log p\left(\bfy |\bfx\right)  }{\partial x_l} }{p(\bfy | \bfx) }
 +
\sigma^2  \frac{\partial \log p(\bfx)}{ \partial x_{l}} 
+ O(\sigma^3),
 \notag\\
 &=
 \sigma^2 
 \expect{ \frac{ \partial }{\partial x_l}
 \log p(\bfx, \bfy)}{p(\bfy | \bfx) }
+ O(\sigma^3),
\label{eq:GradientEstimator}
\end{align}
which completes the proof of Theorem~\ref{thrm:SDAEScore}.

\section{Implementation Details}

We implemented our attacks using \emph{cleverhans} repository \footnote{https://github.com/tensorflow/cleverhans}, code of ALP \citep{kannan2018adversarial}\footnote{https://github.com/tensorflow/models/tree/master/research/adversarial\_logit\_pairing} and also the code of Madry\footnote{https://github.com/MadryLab/mnist\_challenge}. The code was written in Tensorflow \citep{abadi2016tensorflow}
\begin{figure*}[t]
    \centering
    \begin{subfigure}[t]{0.45\textwidth}
        \centering
    \includegraphics[width=\linewidth]{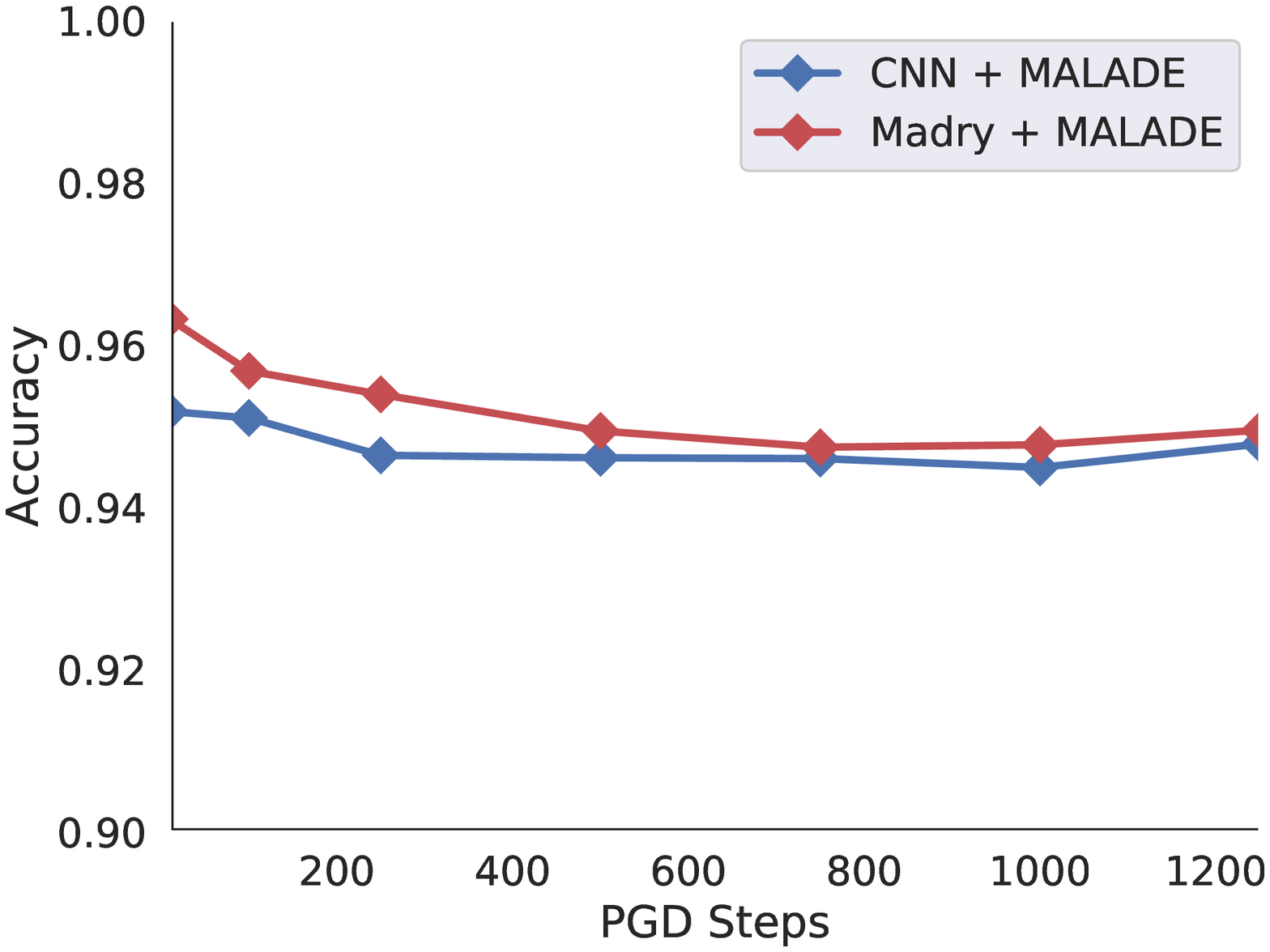}
        \caption{PGD-$L_\infty$, $\varepsilon=0.3$}
        \label{fig:acc_vs_iter_pgd}
    \end{subfigure}
    ~
    \begin{subfigure}[t]{0.45\textwidth}
        \centering
    \includegraphics[width=\linewidth]{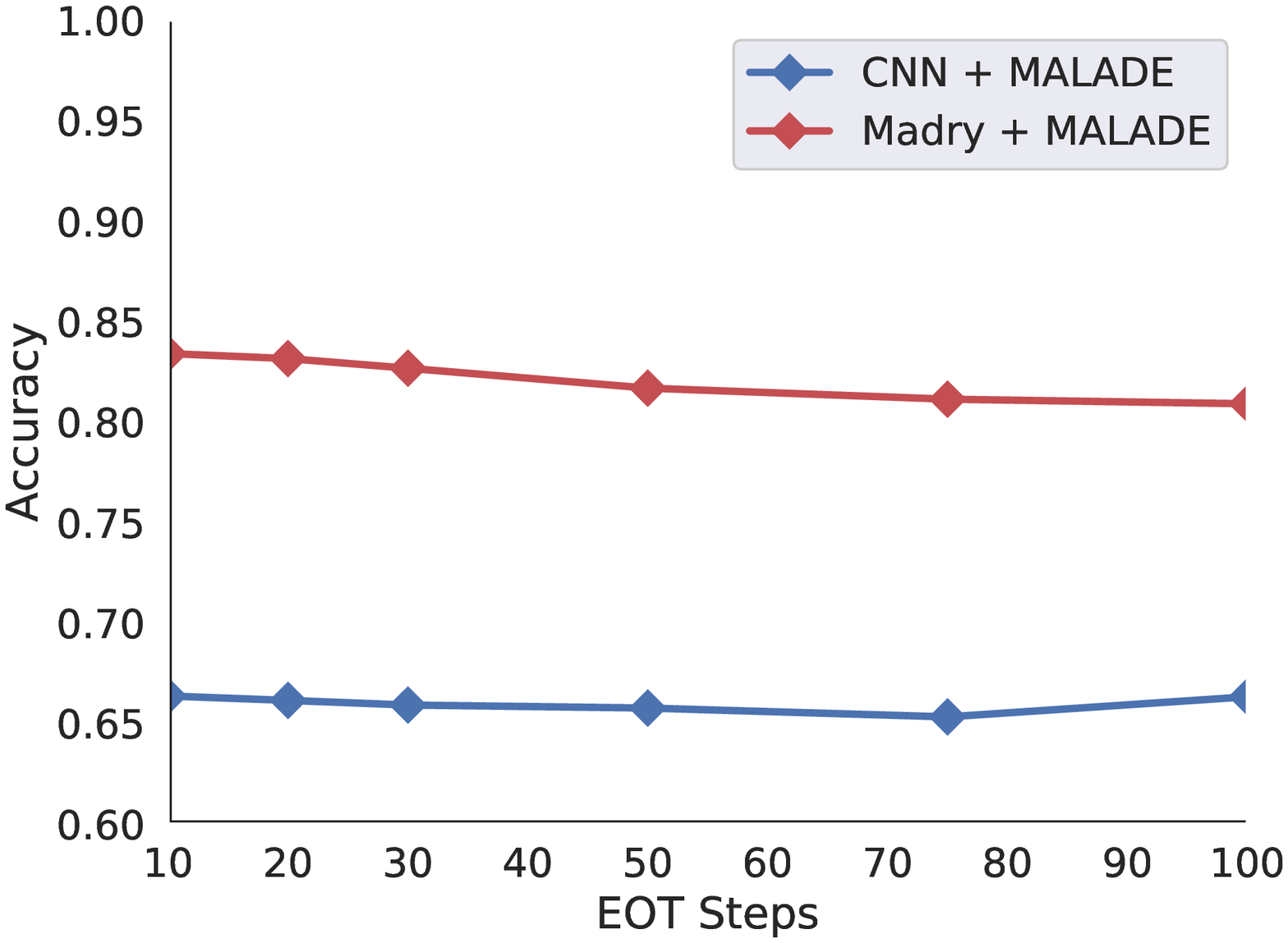}
        \caption{BPDAwEOT-$L_\infty$, $\varepsilon=0.4$}
        \label{fig:acc_vs_iter_bpdaweot}
    \end{subfigure}
\caption{Accuracy is plotted against the number of iterations performed by the attacking method for fixed distortion. }
    \label{fig:acc_vs_iter}
\end{figure*}

\subsection{Hyper-parameter Settings for the Attacks}
\label{sec:App.ParameterSetting}

\subsubsection{PGD}
\label{hyperparams_pgd}
PGD attack was an untargeted attack as it is the simplest strategy for attacking. 
Learning rate of $0.01$ was found to provide for a strong attack while number of iterations was tested for different values as shown in Fig.~\ref{fig:acc_vs_iter_pgd} and fixed at N = $1000$. 
Similar hyper-parameters were used for MIM attack. 

\subsubsection{BPDAwEOT}
\label{hyperparams_bpdaweot}
The attack strategy against \method{} was tested for its effectiveness by varying the number of steps of EOT to be computed.
As shown in Fig.~\ref{fig:acc_vs_iter_bpdaweot} N = $30$ was found to be sufficient for optimal convergence of the attack. 

\subsubsection{CW}
\label{hyperparams_cw}
Hyper-parameters tuned include learning rate = $0.1$ and number of iterations N = $1000$, initial constant $c=100$, binary search step = $1$ and confidence = $0$. The optimizer used here was Adam optimizer.

\subsubsection{EAD}
\label{hyperparams_ead}
Hyper-parameters tuned for attacking \method{} include learning rate = $0.01$ and number of iterations N = $100$, initial constant $c=0.01$, binary search step = $9$ and confidence = $0$. Increasing the number of iterations did not show any increase in the strength of the attack. On the other hand, increasing the number of iterations proved useful for attacking \citep{madry2017towards}, although with all the hyper-parameters being the same. The Adam optimizer was investigated for this attack as recommended by \citep{chen2017ead, sharma2017ead}, however Gradient Descent optimizer proved better in the convergence of the attack.



\subsection{Score Function Estimation}
\subsubsection{Training sDAE}
The score function $\bfnabla_{\bfx}  \log p(\bfx)$ provided by DAE is dependent on the noise $\sigma^2$ added to the input while training the DAE. While too small values for $\sigma^2$ make the score function highly unstable, too large values blur the score. 
The same is true for the score function $\bfnabla_{\bfx}  \log p(\bfx| \bfy)$ provided by \method{}. Here in our experiments on MNIST, we trained the DAE as well as the sDAE with $\sigma^2=0.15$. Such a large noise is beneficial for reliable estimation of the score function \citep{Nguyen}. On the other hand, for CIFAR10 and TinyImagenet datasets, due to the complexity of the dataset, we used $\sigma^2=0.01$. 

\subsubsection{Step Size for Malade}
\label{stepsize}

The score function provided by \method{} drives the generated sample towards high density regions in the data generating distributions. 
With the direction provided by the score function, $\bfalpha$ controls the distance to move agt each step. With large $\bfalpha$, there is possibility of jumping out of the data manifold. While annealing $\bfalpha$ and $\delta^2$ would provides best results as the samples move towards high density region \citep{mala_a, girolami2011riemann}.
In our experiments, we train the sDAE (or DAE) first, followed by searching for good parameters for the $\bfalpha$ and number of steps on the training or validation set. 
This reasonable procedure allows for manually finetuning the step size and number of steps based on the difficulty of the dataset, such that they samples are driven to the nearest high density region of the correct label in polynomial time. 
In the case of adversarial robustness, it also important to select the hyperparameters such that the off-manifold adversarial examples are returned to the data manifold of the correct label in the given number of steps. 
These parameters are then fixed and evaluated on the test set for each of the datasets.

\subsection{Model Architecture}
\label{architectures}
\subsubsection{Classifier and DAE (and sDAE) Architecture on MNIST}

In this appendix, we summarize the architectures of the deep neural networks 
we used in all experiments.
Table~\ref{table:classifier_architectures} gives the architectures of the classifier models while Table~\ref{table:dae_architectures} gives the architecture of the DAE and sDAE models (both have the same architecture). 

\emph{Conv} represents convolution, with the format of Conv(number of output filter maps, kernel size, stride size). \emph{Linear} represents a fully connected layer with the format of Linear(number of output neurons). \emph{Relu} is a rectified linear unit while \emph{Tanh} is a hyberbolic tangent function. \emph{Softmax} is a logistic function which squashes the input tensor to real values of range [0,1] and adds up to 1. \emph{Conv\_Transpose} is transpose of the convolution operation, sometimes called deconvolution, with the format of Conv\_Transpose(number of output filter maps, kernel size, stride size).

\label{classifiers}
\begin{table}[h]
  \caption{Architectures of Classifier}
  \label{table:classifier_architectures}
  \centering{}
  \begin{tabular}{l c  }
    \toprule
    CNN	 \\
    \midrule
      	 \\
    Conv($128, 3\times3, 1$)  \\
    Relu() 					   \\
    Conv($64, 5\times5, 2$)   \\
    Relu()                    \\
    Linear($128$)             \\
    Relu()		                \\
    Linear($10$)                \\
    Softmax() 	                 \\

  \end{tabular}
  \vspace{1mm}	
 \end{table}


\begin{table}[h]
  \caption{Architecture of DAE and sDAE.}
  \label{table:dae_architectures}
  \centering{}
  \begin{tabular}{l c c}
    \toprule
    & DAE, sDAE 	 \\
    \midrule
      	 \\
    		&Conv($10, 5\times5, 2$) \\
   Encoder  &Tanh() 					\\
    		&Conv($25, 5\times5, 2$) \\
    		&Tanh() 					 \\
            \midrule
    		&Conv\_Transpose($10, 9\times9, 2$) \\
    		&Tanh()  				\\
   Decoder  &Conv\_Transpose($1, 1\times1, 2$) \\
    		&Tanh() \\
    		&Conv($1, 5\times5, 1$) \\
    		&Tanh() 					 \\
    
  \end{tabular}
  \vspace{1mm}	
 \end{table}

\subsubsection{CIFAR10}
\label{architecutre_cifar10}
The CNN architecture for CIFAR10 was borrowed from Tensorflow library\footnote{https://github.com/tensorflow/models/tree/master/official/resnet}. 
The model of Madry for CIFAR10 dataset is as implemented by the authors in \cite{madry2017towards}. 

\subsubsection{TinyImagenet}
\label{architecutre_tinyimagenet}
The architecture and the pretrained models of CNN as well as ALP is as implemented by the authors\footnote{https://github.com/tensorflow/models/tree/master/research/adversarial\_logit\_pairing}.

\newpage


\section{Effect of EOT}

\begin{figure}[h]
    
    \centering
    \includegraphics[width=0.6\linewidth]{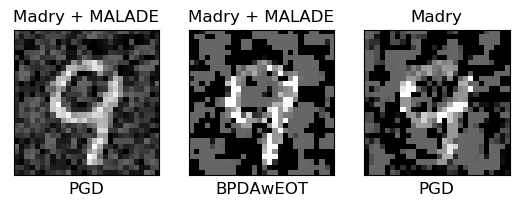}
    
\caption{The adversarial examples shown here correspond to - PGD-$L_\infty$ attack for Madry + \method{}, BPDAwEOT for Madry + \method{} and finally PGD-$L_\infty$ for Madry. The maximum perturbation allowed in all of these settings was $\varepsilon=0.4$. The effect of gradient obfuscation is visible in the first image while BPDAwEOT is effective in preventing this phenomenon. 
}
\label{fig:grad_obfusc_mnist}
\end{figure}

In order to eliminate random effect, and obtain useful gradient information for the attacker, we formulated BPDAwEOT against Madry + \method{} and found this to be the strongest attack. 
Fig.~\ref{fig:grad_obfusc_mnist} shows the 
adversarial perturbations of PGD on Madry + \method{} results in perturbations that appear to be blobs throughout the image. 
Adversarial images from BPDAwEOT on the other hand are similar to the perturbations crafted by the attacker for PGD on Madry alone -- indicating that the phenomenon of gradients obfuscation is prevented. 
Nevertheless, the adversarial samples against \method{} are not as effective as against Madry, as shown in Fig.~\ref{fig:attack_compare}.
This is because the inherent randomness of \method{} prevents the attack from stably aligning the sample to a targeted untrained spot.

\newpage

\section{Adversarial Samples}
\label{images_mnist}

\begin{figure*}[!ht]
    \centering
    \begin{subfigure}[t]{0.3\textwidth}
        \centering
    \includegraphics[width=\linewidth]{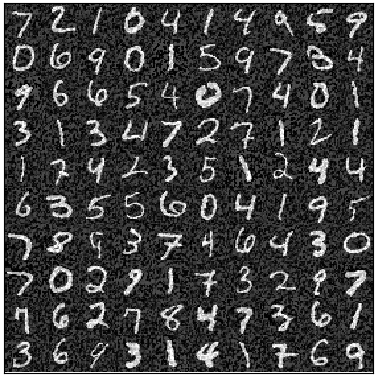}
        \caption{$\varepsilon =  0.3$ }
    \end{subfigure}
    ~
    \begin{subfigure}[t]{0.3\textwidth}
        \centering
    \includegraphics[width=\linewidth]{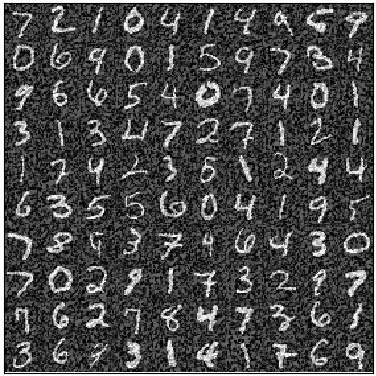}
        \caption{$\varepsilon =  0.4$ }
    \end{subfigure}
\caption{Sample adversarial images crafted by a PGD attack with $L_\infty$ norm are shown here. }
    \label{fig:appendix_pgd}
\end{figure*}

\begin{figure*}[!ht]
    \centering
    \begin{subfigure}[t]{0.3\textwidth}
        \centering
    \includegraphics[width=\linewidth]{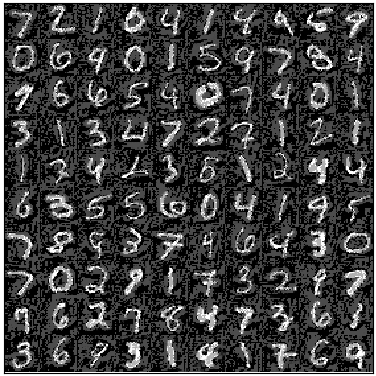}
        \caption{$\varepsilon =  0.3$}
    \end{subfigure}
    ~
    \begin{subfigure}[t]{0.3\textwidth}
        \centering
    \includegraphics[width=\linewidth]{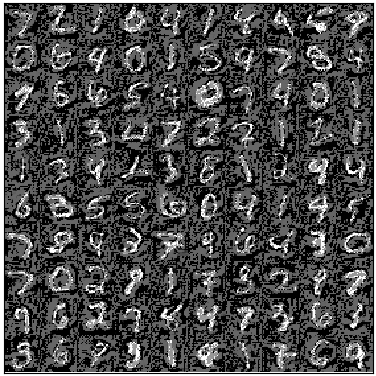}
        \caption{$\varepsilon =  0.4$}
    \end{subfigure}
\caption{Sample adversarial images crafted by R+PGD attack with $L_\infty$ norm are shown here. }
    \label{fig:appendix_pgdrec}
\end{figure*}

\begin{figure*}[!ht]
    \centering
    \begin{subfigure}[t]{0.3\textwidth}
        \centering
    \includegraphics[width=\linewidth]{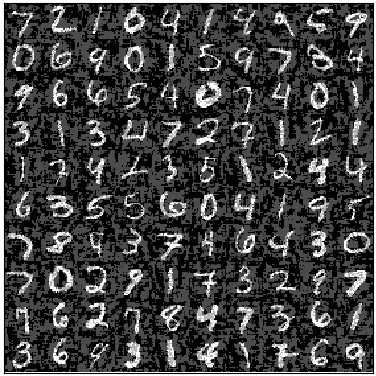}
        \caption{$\varepsilon =  0.3$ }
    \end{subfigure}
    ~
    \begin{subfigure}[t]{0.3\textwidth}
        \centering
    \includegraphics[width=\linewidth]{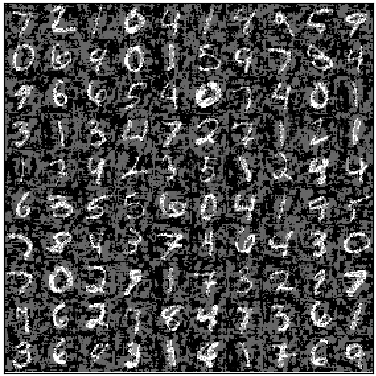}
        \caption{$\varepsilon =  0.4$ }
    \end{subfigure}
\caption{Sample adversarial images crafted by a BPDA attack with $L_\infty$ norm are shown here. }
    \label{fig:appendix_bpda}
\end{figure*}

\begin{figure*}[!ht]
    \centering
    \begin{subfigure}[t]{0.3\textwidth}
        \centering
    \includegraphics[width=\linewidth]{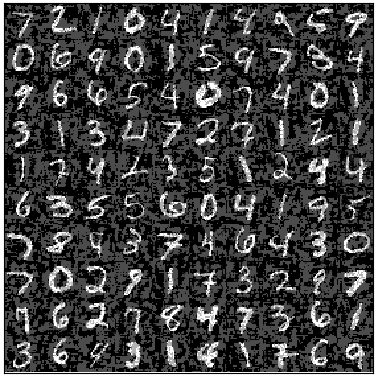}
        \caption{$\varepsilon =  0.3$ }
    \end{subfigure}
    ~
    \begin{subfigure}[t]{0.3\textwidth}
        \centering
    \includegraphics[width=\linewidth]{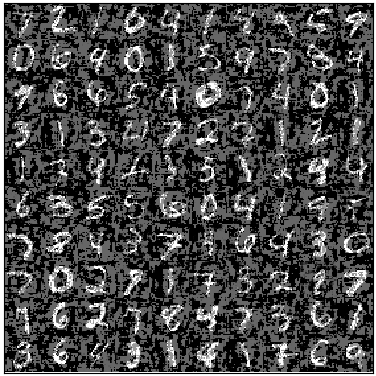}
        \caption{$\varepsilon =  0.4$ }
    \end{subfigure}
\caption{Sample adversarial images crafted by a BPDAwEOT attack with $L_\infty$ norm are shown here. }
    \label{fig:appendix_bpdaweot}
\end{figure*}

\begin{figure*}[!ht]
    \centering
    \begin{subfigure}[t]{0.3\textwidth}
        \centering
    \includegraphics[width=\linewidth]{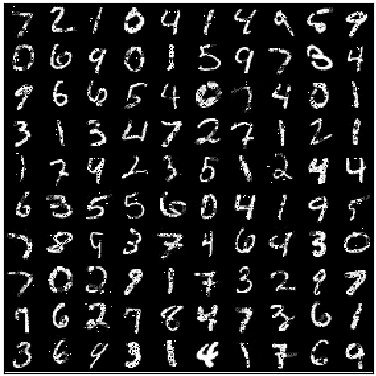}
        \caption{Adversarial Examples against Madry with $\varepsilon_{L_\infty} = 0.93, \varepsilon_{L_1} = 125.39, \varepsilon_{L_2} = 8.18$}
    \end{subfigure}
    ~
    \begin{subfigure}[t]{0.3\textwidth}
        \centering
    \includegraphics[width=\linewidth]{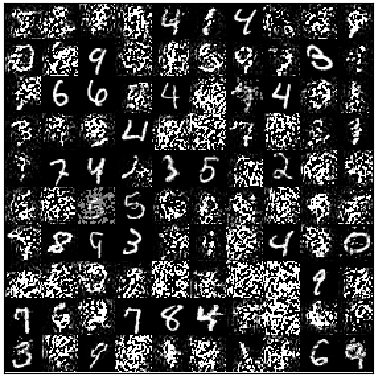}
        \caption{Adversarial Examples against Madry + \method{} with $\varepsilon_{L_\infty} = 0.83, \varepsilon_{L_1} = 22.64, \varepsilon_{L_2} = 3.55$}
    \end{subfigure}
\caption{Sample adversarial images crafted by EAD attack are shown here. The attack fails to converge for many images despite our best effort to finetune the algorithm.}
    \label{fig:appendix_ead}
\end{figure*}

\begin{figure*}[!ht]
    \centering
    \begin{subfigure}[t]{0.3\textwidth}
        \centering
    \includegraphics[width=\linewidth]{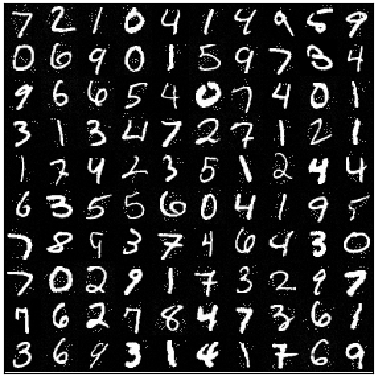}
        \caption{Adversarial Examples against Madry with $\varepsilon_{L_\infty} = 0.56, \varepsilon_{L_1} = 11.36, \varepsilon_{L_2} = 1.38$}
    \end{subfigure}
    ~
    \begin{subfigure}[t]{0.3\textwidth}
        \centering
    \includegraphics[width=\linewidth]{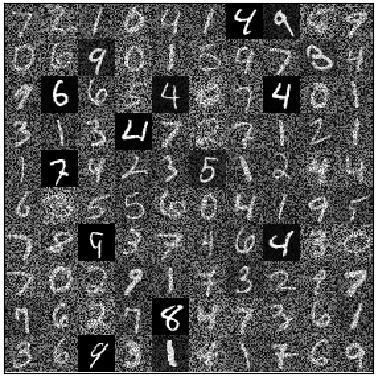}
        \caption{Adversarial Examples against Madry + \method{} with $\varepsilon_{L_\infty} = 0.54, \varepsilon_{L_1} = 200.46, \varepsilon_{L_2} = 8.34$} 
    \end{subfigure}
\caption{Sample adversarial images crafted by a Boundary attack are shown here. Some images fail to become adversarial during the initialization of the algorithm with random uniform noise and hence are retained as the original image. }
    \label{fig:appendix_boundary}
\end{figure*}


\begin{figure*}[!ht]
    \centering
    \begin{subfigure}[t]{0.5\textwidth}
        \centering
    \includegraphics[width=\linewidth]{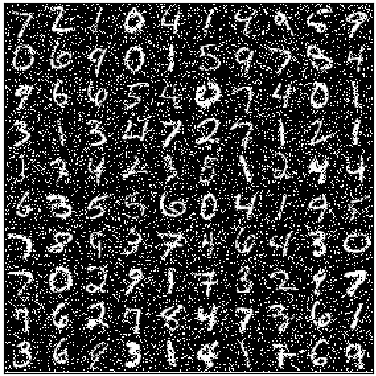}
        \caption{Images corrupted by Salt and Pepper Noise Attack with $\varepsilon_{L_\infty} = 0.99, \varepsilon_{L_1} = 98.02, \varepsilon_{L_2} = 9.70$}
    \end{subfigure}
\caption{Sample adversarial images crafted by adding salt and pepper noise 
are shown here. The mean of the magnitude of the perturbations over the entire test dataset for each distance measure is given below each image. }
    \label{fig:appendix_saltnpepper}
\end{figure*}

\begin{figure*}[!ht]
    \centering
    \begin{subfigure}[t]{0.45\textwidth}
        \centering
    \includegraphics[width=\linewidth]{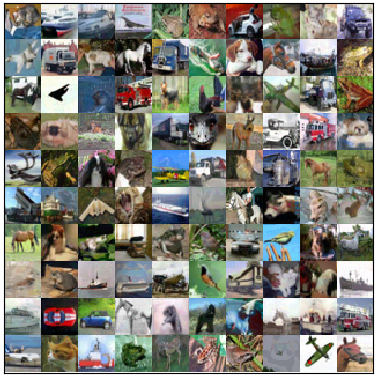}
        \caption{$\varepsilon =  8.0$ }
    \end{subfigure}
    ~
    \begin{subfigure}[t]{0.45\textwidth}
        \centering
    \includegraphics[width=\linewidth]{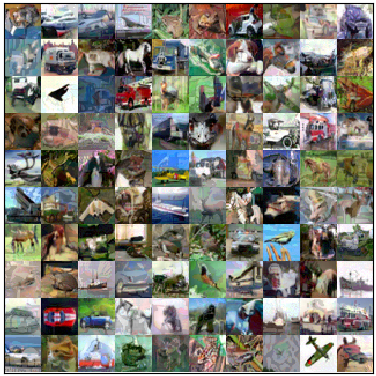}
        \caption{$\varepsilon =  16.0$ }
    \end{subfigure}
\caption{Sample adversarial images crafted by PGD $L_\infty$ attack against Madry + \method{} for the CIFAR10 dataset are shown here. }
    \label{fig:appendix_cifar10}
\end{figure*}

\begin{figure*}[!ht]
    \centering
    \begin{subfigure}[t]{0.75\textwidth}
        \centering
    \includegraphics[width=\linewidth]{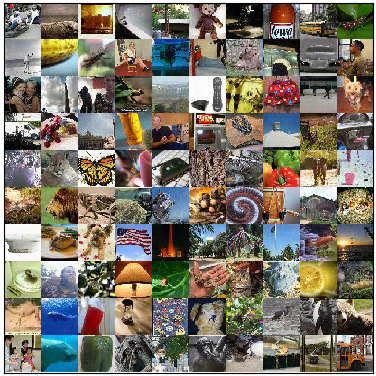}
        \caption{$\varepsilon =  8.0$ }
    \end{subfigure}
    ~
    \begin{subfigure}[t]{0.75\textwidth}
        \centering
    \includegraphics[width=\linewidth]{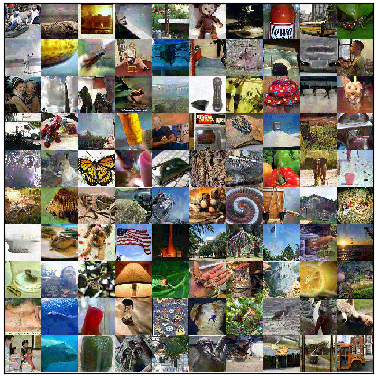}
        \caption{$\varepsilon =  16.0$ }
    \end{subfigure}
\caption{Sample adversarial images crafted by PGD $L_\infty$ attack against ALP + \method{} for the TinyImagenet dataset are shown here. }
    \label{fig:appendix_tinyimagenet}
\end{figure*}

\begin{figure*}[!h]
    \centering
    \begin{subfigure}[t]{0.7\textwidth}
        \centering
    \includegraphics[width=\linewidth]{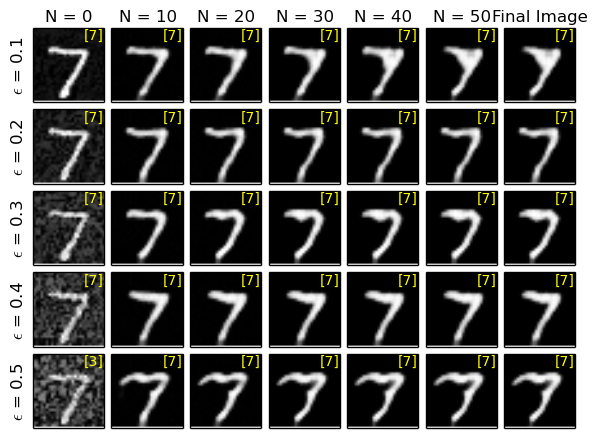}
        \caption{}
    \end{subfigure}
    ~
    \begin{subfigure}[t]{0.7\textwidth}
        \centering
    \includegraphics[width=\linewidth]{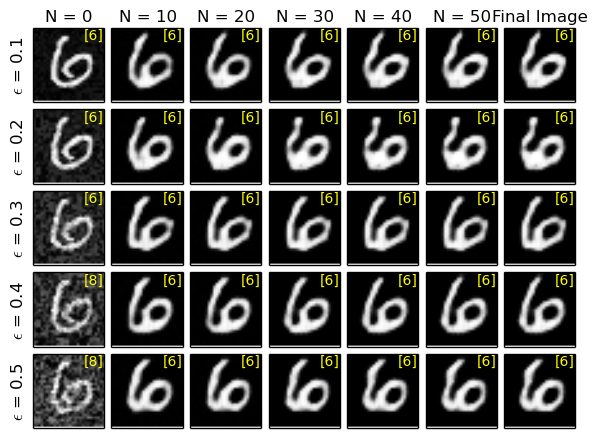}
        \caption{}
    \end{subfigure}
\caption{Sample images for the \method{} algorithm against a PGD attack with $L_\infty$ norm are shown here. The rows indicate the norm of the perturbation used by the attacked while the columns indicate the intermittent steps taken by \method{} to defend the attack. 
The classifier's decision is displayed in yellow in the top right corner of each image. Due to gradient obfuscation, the adversarial sampels are not very strong and hence \method{} is very robust here.}
    \label{fig:appendix_pgd_linf}
\end{figure*}

\begin{figure*}[!h]
    \centering
    \begin{subfigure}[t]{0.7\textwidth}
        \centering
    \includegraphics[width=\linewidth]{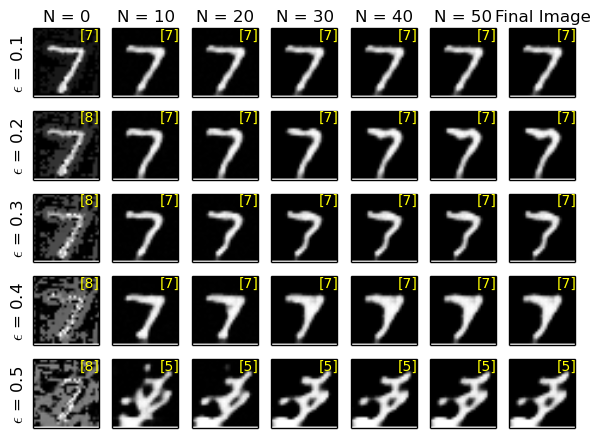}
        \caption{}
    \end{subfigure}
    ~
    \begin{subfigure}[t]{0.7\textwidth}
        \centering
    \includegraphics[width=\linewidth]{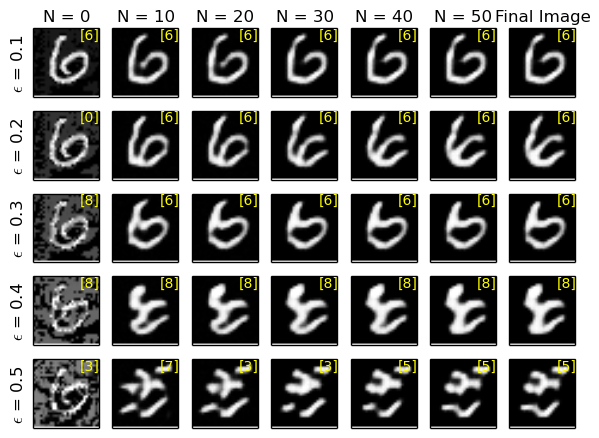}
        \caption{}
    \end{subfigure}
\caption{Sample images for the \method{} algorithm against a BPDAwEOT attack with $L_\infty$ norm are shown here. The rows indicate the norm of the perturbation used by the attacked while the columns indicate the intermittent steps taken by \method{} to defend the attack. 
The classifier's decision is displayed in yellow in the top right corner of each image. Since the gradients are computed only until the input to the classifier, there is no gradient obfuscation and hence the attack is strong.}
    \label{fig:appendix_bpda}
\end{figure*}

\newpage~\newpage
\begin{figure*}[!ht]
    \centering
    \begin{subfigure}[t]{0.7\textwidth}
        \centering
    \includegraphics[width=\linewidth]{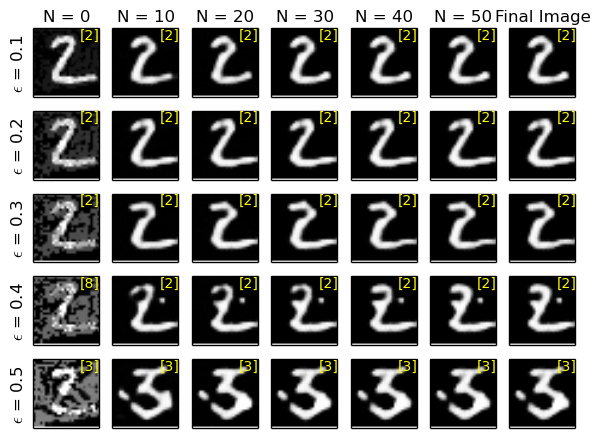}
        \caption{}
    \end{subfigure}
    ~
    \begin{subfigure}[t]{0.7\textwidth}
        \centering
    \includegraphics[width=\linewidth]{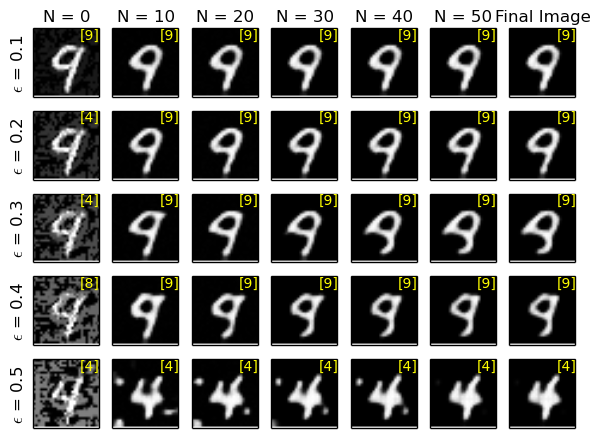}
        \caption{}
    \end{subfigure}
\caption{Sample images for the \method{} algorithm against a BPDAwEOT attack with $L_\infty$ norm are shown here. The rows indicate the norm of the perturbation used by the attacked while the columns indicate the intermittent steps taken by \method{} to defend the attack. 
The classifier's decision is displayed in yellow in the top right corner of each image. }
    \label{fig:appendix_pgd_linf_success}
\end{figure*}

\newpage
\begin{figure*}[!ht]
    \centering
    \begin{subfigure}[t]{0.7\textwidth}
        \centering
    \includegraphics[width=\linewidth]{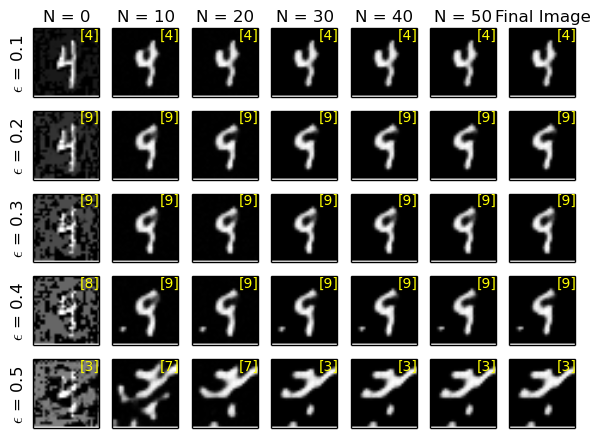}
        \caption{}
    \end{subfigure}
    ~
    \begin{subfigure}[t]{0.7\textwidth}
        \centering
    \includegraphics[width=\linewidth]{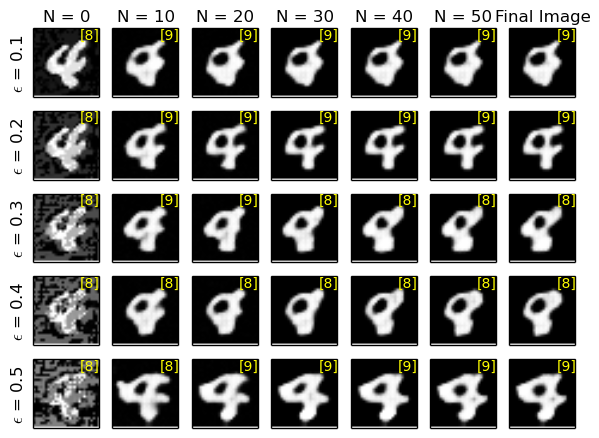}
        \caption{}
    \end{subfigure}
\caption{Sample images for the \method{} algorithm against a BPDAwEOT attack with $L_\infty$ norm are shown here. The rows indicate the norm of the perturbation used by the attacked while the columns indicate the intermittent steps taken by \method{} to defend the attack. 
The classifier's decision is displayed in yellow in the top right corner of each image. }
    \label{fig:appendix_pgd_linf_failure}
\end{figure*}
\newpage

\end{document}